\documentclass[acmsmall]{acmart}
\usepackage{multirow}
\usepackage{amsmath}
\usepackage{amssymb}
\usepackage{booktabs}
\usepackage{tabu}
\usepackage{color}

\definecolor{MyBlue}{RGB}{50,85,164} 
\definecolor{DarkGreen}{RGB}{85, 80, 37}
\definecolor{LightGreen}{RGB}{187, 197, 146}
\definecolor{BrightGreen}{RGB}{181, 189, 0}
\definecolor{MidGreen}{RGB}{143, 153, 62}
\definecolor{green}{rgb}{0.13, 0.55, 0.13}
\definecolor{red}{rgb}{0.86, 0.08, 0.24}

\AtBeginDocument{%
  \providecommand\BibTeX{{%
    \normalfont B\kern-0.5em{\scshape i\kern-0.25em b}\kern-0.8em\TeX}}}

\begin{document}

\title{Will You Ever Become Popular? Learning to Predict Virality of Dance Clips}

\author{Jiahao Wang}
\email{jhwang@buaa.edu.cn}

\author{Yunhong Wang}
\email{yhwang@buaa.edu.cn}

\author{Nina Weng}
\email{wengnn@buaa.edu.cn}

\author{Tianrui Chai}
\email{trchai@buaa.edu.cn}

\author{Annan Li}
\authornote{Corresponding author.}
\email{liannan@buaa.edu.cn}
\affiliation{%
  \institution{State Key Laboratory of Virtual Reality Technology and System, Beihang University}
  \city{Beijing}
  \country{China}}

\author{Faxi Zhang}
\email{micahzhang@tencent.com}

\author{Sansi Yu}
\email{mionyu@tencent.com}
\affiliation{%
  \institution{Tencent}
  \city{Shenzhen}
  \country{China}
}

\renewcommand{\shortauthors}{Jiahao Wang, et al.}

\begin{abstract}
Dance challenges are going viral in video communities like TikTok nowadays. 
Once a challenge becomes popular, thousands of short-form videos will be uploaded in merely a couple of days.
Therefore, virality prediction from dance challenges is of great commercial value and has a wide range of applications, such as smart recommendation and popularity promotion.
In this paper, a novel multi-modal framework which integrates skeletal, holistic appearance, facial and scenic cues is proposed for comprehensive dance virality prediction.
To model body movements, we propose a pyramidal skeleton graph convolutional network (PSGCN) which hierarchically refines spatio-temporal skeleton graphs. 
Meanwhile, we introduce a relational temporal convolutional network (RTCN) to exploit appearance dynamics with non-local temporal relations.
An attentive fusion approach is finally proposed to adaptively aggregate predictions from different modalities.
To validate our method, we introduce a large-scale viral dance video (VDV) dataset, which contains over 4,000 dance clips of eight viral dance challenges.
%
Extensive experiments on the VDV dataset well demonstrate the effectiveness of our approach. 
Furthermore, we show that short video applications like multi-dimensional recommendation and action feedback can be derived from our model.
\end{abstract}


\setcopyright{acmcopyright}
\acmJournal{TOMM}
\acmYear{2021} \acmVolume{1} \acmNumber{1} \acmArticle{1} \acmMonth{1} \acmPrice{15.00}\acmDOI{10.1145/3477533}


\maketitle

\section{Introduction}
\label{sec:intro}

\begin{figure*}[t]
\begin{center}
   \includegraphics[width=1.0\textwidth]{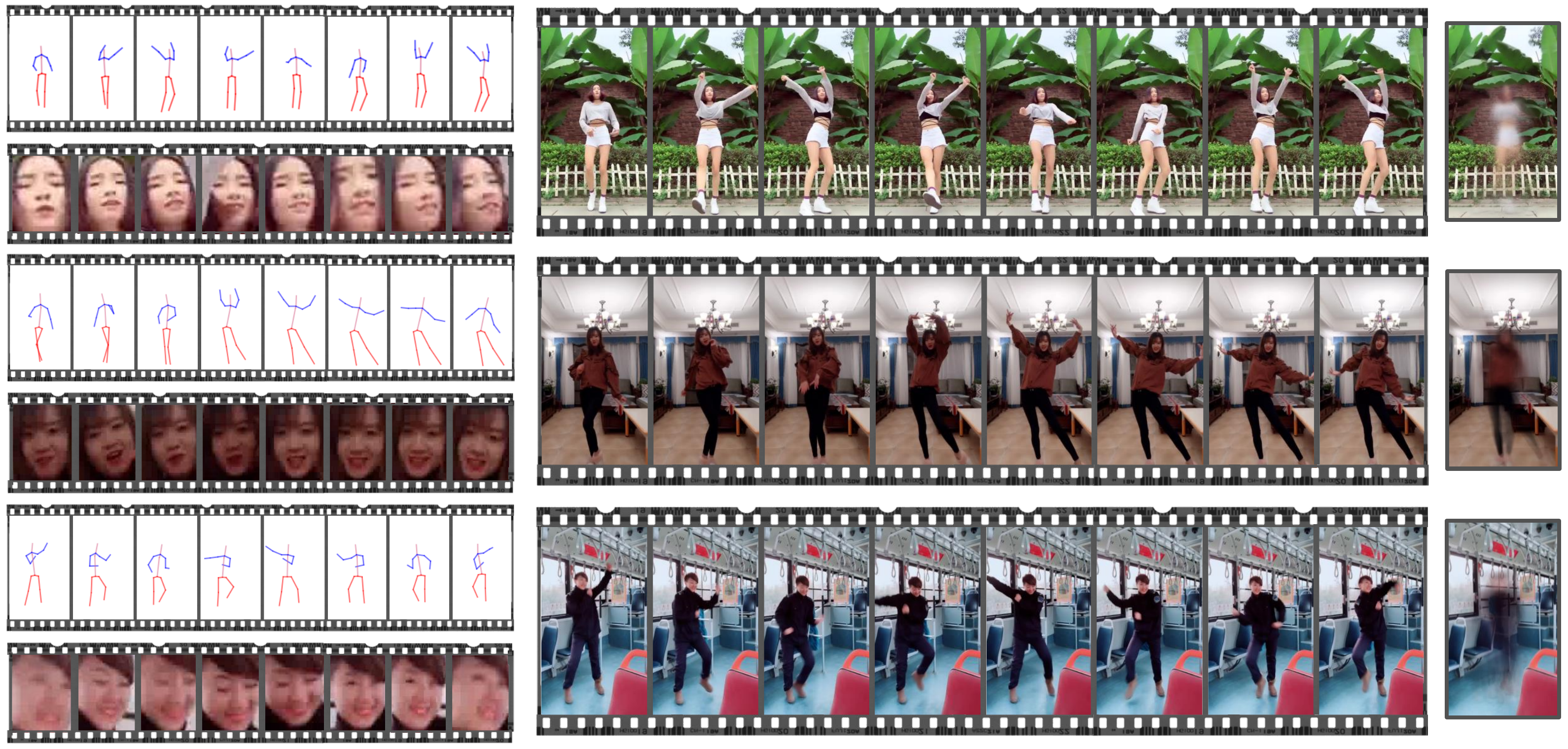}
\end{center}
   \vspace{-5mm}
   \caption{Viral dance clips usually have various visual features, such as exaggerated body movements, attractive facial expressions, stylish backgrounds, etc. This figure shows three examples from our VDV dataset and corresponding skeletal (top left), facial (bottom left), scenic (right) features.}
   \vspace{-5mm}
\label{fig:intro}
\end{figure*}

Music-centric short videos, such as dance challenge, short lip-sync and spoofs, are booming on social media. Take TikTok (a.k.a. Douyin in China) as an example, it is recorded to have 689 million monthly active users worldwide by early 2021~\cite{Mohsin2021tiktok}. For the reason of limited network bandwidth and fragmented reading behavior, the duration of video is usually set to less than one minute, typically 3-15 seconds. Since such short videos are purposely created and highly topic-driven, it results in a huge number of well-formated, richly annotated videos and creates very specific demands of video content analysis. Dance challenge is one of the hottest topics of the short video community, in which people perform dance with the same background music and compete with each other for higher popularity. Once a challenge becomes popular, thousands of challengers will upload their performance. However, most of the uploads are parodies and only a fraction of them can go viral. Therefore, automatically predicting the potential of virality for a new upload becomes an important issue. For short video platforms, the resulting implications can be used for content recommendation, traffic pool management, video retrieval and advertising. For end users, prompts and feedback can be acquired to raise their popularity.

Over the past few years, plenty of works have been done to predict the popularity of online videos, in which samples from YouTube~\cite{pinto2013using,trzcinski2017predicting} and Facebook~\cite{bielski2018pay, bielski2018understanding} with miscellaneous content have been investigated. Researchers utilize multi-modal data including the early evolution pattern and social media, visual, acoustic, textual features~\cite{pinto2013using,trzcinski2017predicting,bielski2018pay,chen2016micro} to make popularity prediction. Meanwhile, in the domain of human-centric performance assessment, models are developed to predict performance scores of Olympic events~\cite{pirsiavash2014assessing,parmar2017learning,xiang2018s3d,parmar2019and,pan2019action}, surgical skill~\cite{gao2014jhu,zia2015automated} and patient rehabilitation~\cite{parmar2016measuring,tao2016comparative}. 
These works usually take single modality data (e.g. human skeleton, video frames) as the assessment basis.

\begin{figure*}[t]
\begin{center}
   \includegraphics[width=1.0\textwidth]{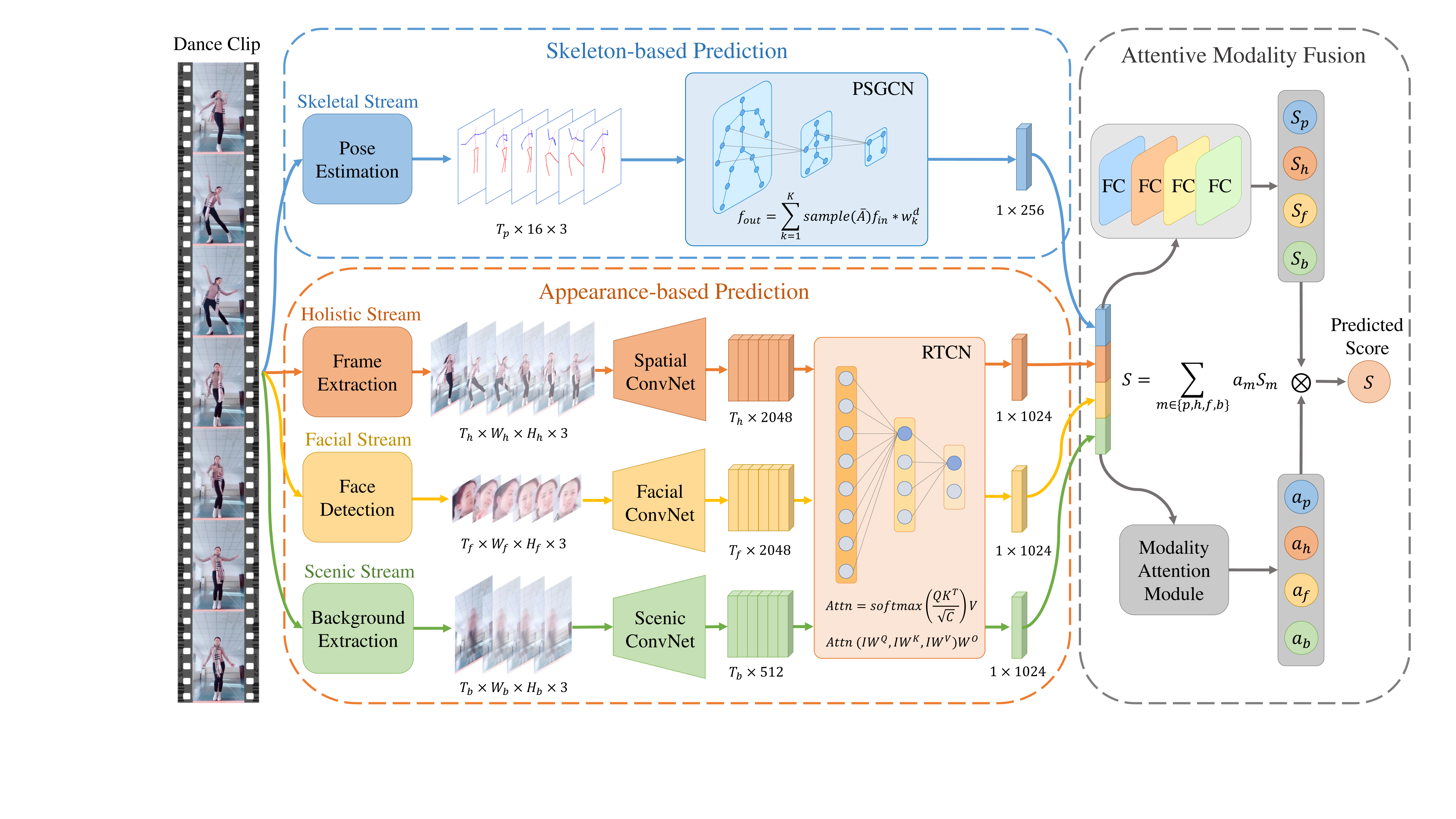}
\end{center}
   \vspace{-3mm}
   \caption{Overview of our multi-modal framework. Given an input dance clip, skeletal, holistic, facial and scenic features are extracted by state-of-the-art deep models. For skeleton-based virality prediction, we introduce PSGCN to exploit human skeleton dynamics. For appearance-based prediction, holistic, facial and scenic features are explored with RTCN models. 
   An attentive modality fusion method is finally proposed to aggregate multi-modal predictions.
   Fully-connected (FC) layers are added at the end of each stream to generate prediction scores, which are then fused by weights from the modality attention module.
   Sizes of intermediate features are given at the bottom. Best viewed in color.}
   \vspace{-5mm}
\label{fig:framework}
\end{figure*}

In this paper, we focus on predicting virality of dance clips with visual cues. Unlike existing video popularity prediction works that deal with universal videos with miscellaneous content, we only pay attention to single-person dance clips within exclusive dance challenges. Our virality score is based on several popularity indicators and reflects the public appreciation. As illustrated in Fig.~\ref{fig:intro}, the virality of a dance clip can be influenced by various factors, such as body movements, facial expressions, backgrounds, etc. 
To comprehensively predict the virality, we propose a multi-modal prediction framework that consists of a skeleton-based stream and three appearance-based streams, 
accounting for body movements and appearance cues of multiple aspects, respectively. The overview of our framework is presented in Fig.~\ref{fig:framework}. 

For skeleton-based virality prediction, we introduce a pyramidal skeleton graph convolutional network (PSGCN), which hierarchically exploits human skeleton dynamics with spatial-temporal graph convolutions. In order to capture robust virality information across different dance challenges, we design a graph down-sampling module to gradually refine the spatio-temporal graphs, forming a pyramidal network architecture.
For appearance-based virality prediction, we employ state-of-the-art deep architectures~\cite{wang2016temporal,zhang2016joint,cao2018vggface2,zhou2017places} to extract multi-modal features with holistic, facial and scenic information. An effective and efficient relational temporal convolutional network (RTCN) is proposed to extract appearance dynamics from frame-level features. 
By inserting the devised relational pooling blocks between temporal convolutions, we explore non-local temporal relations via multi-head self-attentions.
To aggregate predictions from different modalities, we adopt a late fusion~\cite{karpathy2014large,wang2016temporal} scheme facilitated by a data-dependent modality attention mechanism, which employs a modality attention module to adaptively learn attention weights for different streams according to the input video.

Finally, to validate our approach, we present a large-scale viral dance video (VDV) dataset, which contains 4,292 dance clips sampled from eight representative challenges with a total length of 17.1 hours. To facilitate the development of multi-modal solutions, we also release the data in skeletal, facial and scenic modalities. Extensive experiments on VDV dataset demonstrate the superiority of the proposed approach compared with state-of-the-art methods. 

Moreover, we show that applications for both short video platforms and end users, such as multi-dimensional video recommendation and action feedback, can be derived from our model.
Extension experiments on UNLV-Sports~\cite{parmar2017learning} dataset indicate that our multi-modal framework is beneficial to other human-centric performance assessment tasks as well. 

In summary, our main contributions are three-fold:
\begin{itemize}
\item We first study virality prediction from dance challenges using visual cues, which has great commercial value. To facilitate the research, we release VDV dataset, a large-scale multi-modal dance virality prediction benchmark.
\item A multi-modal framework modeling both body movements and appearance dynamics is developed. For skeleton-based prediction, we devise a pyramidal skeleton graph convolutional network (PSGCN). For appearance-based prediction, relational temporal convolutional networks (RTCN) are proposed. An attentive modality fusion approach is introduced to aggregate predictions from multiple streams.
\item Our model achieves the state-of-the-art performance on VDV dataset. Real-world short video applications like multi-dimensional video recommendation and action feedback can be derived from our model. Extension experiments indicate that our multi-modal framework has potential in other assessment tasks like sports rating.
\end{itemize}

The rest of this paper is structured as follows. Section~\ref{sec:related_work} reviews related works and their relations with this paper. Section~\ref{sec:data_collection} introduces the proposed VDV dataset. Section~\ref{sec:approach} presents details of our multi-modal prediction framework. Section~\ref{sec:experiments} shows experimental evaluations of our approach and the competitor methods. Finally, Section~\ref{sec:conclusion} concludes the paper.

\section{Related Work}
\label{sec:related_work}

\noindent\textbf{Online video popularity prediction.} Researchers from multimedia and data mining communities have done a considerable number of works to predict the popularity of online videos. Most of these works intend to predict the popularity of videos from online video or social media websites, such as YouTube~\cite{pinto2013using,trzcinski2017predicting,wongsuparatkul2020view}, Vine~\cite{chen2016micro,nie2019multimodal,su2020predicting,xie2020multimodal},
Facebook~\cite{bielski2018pay,bielski2018understanding}, Kuaishou~\cite{zhang2020graphinf}, etc. Multi-modal data including popularity evolution pattern and social media, visual, acoustic, textual, geography
features are exploited for prediction. In~\cite{pinto2013using}, view counts of YouTube videos at
early stages are used to predict those in the future. Chen et al.~\cite{chen2016micro} propose to predict the popularity of micro-videos from Vine using a transductive model, in which social, visual, acoustic and textual features are taken as the input. 
Following this work, methods like variational encoder-decoder~\cite{xie2020multimodal} and feature-discrimination transductive model~\cite{su2020predicting} are explored to better capture popularity-related information from the same features.
Visual cues are combined with early evolution patterns in~\cite{trzcinski2017predicting} to train an support vector regressor (SVR) for popularity prediction.
In~\cite{bielski2018pay}, a spatial-attentive network is proposed to explore the relationship between visual appearance and video virality. 
A graph convolution based model is introduced in~\cite{zhang2020graphinf} to predict video popularity from geography information of users.

In comparison, our work focuses on single-person dance clips in short video apps and aims to predict the virality with multi-modal visual cues. As later introduced in Section~\ref{sec:data_collection}, the videos in VDV dataset are carefully selected to ensure that the virality mainly depends on the video content. Unlike most of the video popularity prediction works that utilize non-visual data (e.g. early evolution pattern, social media data, etc.) for prediction, we only adopt visual features since we concentrate on the virality of dance performance itself, rather than other irrelevant information.

\noindent\textbf{Human-centric performance assessment.} Human-centric performance assessment, which aims to predict the performance score of human individuals in certain kinds of activities, is attracting increasing attention due to its broad applications in multiple areas. Some researchers focus on assessing the performance of athletes in Olympic events, such as diving, figure skating and gymnastic vault~\cite{pirsiavash2014assessing,parmar2017learning,xiang2018s3d,parmar2019and,pan2019action,tang2020uncertainty}, some target on skill determination in daily activities~\cite{doughty2018s,doughty2019pros}, others address medical issues like surgical skill~\cite{gao2014jhu,zia2015automated,gao2020asymmetric} and patient rehabilitation~\cite{parmar2016measuring,tao2016comparative}. In comparison, little attention is paid to entertainment videos like viral dance clips. In terms of data modality, most works only utilize single modality data for assessment, such as human skeleton~\cite{pirsiavash2014assessing,venkataraman2015dynamical,parmar2016measuring}, video frames~\cite{parmar2017learning,xiang2018s3d,parmar2019and,parmar2019action}. Pirsiavash et al.~\cite{pirsiavash2014assessing} use the discrete cosine transform (DCT) to represent human skeleton features and train SVR models to predict performance scores of Olympic events. Later works also explore other machine learning methods including the hidden Markov model~\cite{tao2016comparative} and boosted decision trees~\cite{parmar2016measuring} to directly predict the performance score from human skeleton features. Due to the difficulty in estimating atypical body postures~\cite{parmar2017learning}, recent works tend to utilize visual features directly extracted from video frames instead. In~\cite{parmar2017learning,parmar2019action}, C3D models~\cite{tran2015learning} are used to extract spatio-temporal features from input videos. Models like the long short-term memory (LSTM) and SVR are used to regress the performance score. A segment-based pseudo-3D residual network (P3D)~\cite{qiu2017learning} is proposed in~\cite{xiang2018s3d} to extract features at segment-level for better performance. Later works like~\cite{gao2020asymmetric,tang2020uncertainty} employ I3D~\cite{carreira2017quo} model for segmental feature extraction. In~\cite{tang2020uncertainty}, an uncertainty-aware score distribution learning (USDL) approach is proposed to handle the score ambiguity caused by multiple judges in sports events.

\begin{figure*}
\begin{center}
   \includegraphics[width=0.95\textwidth]{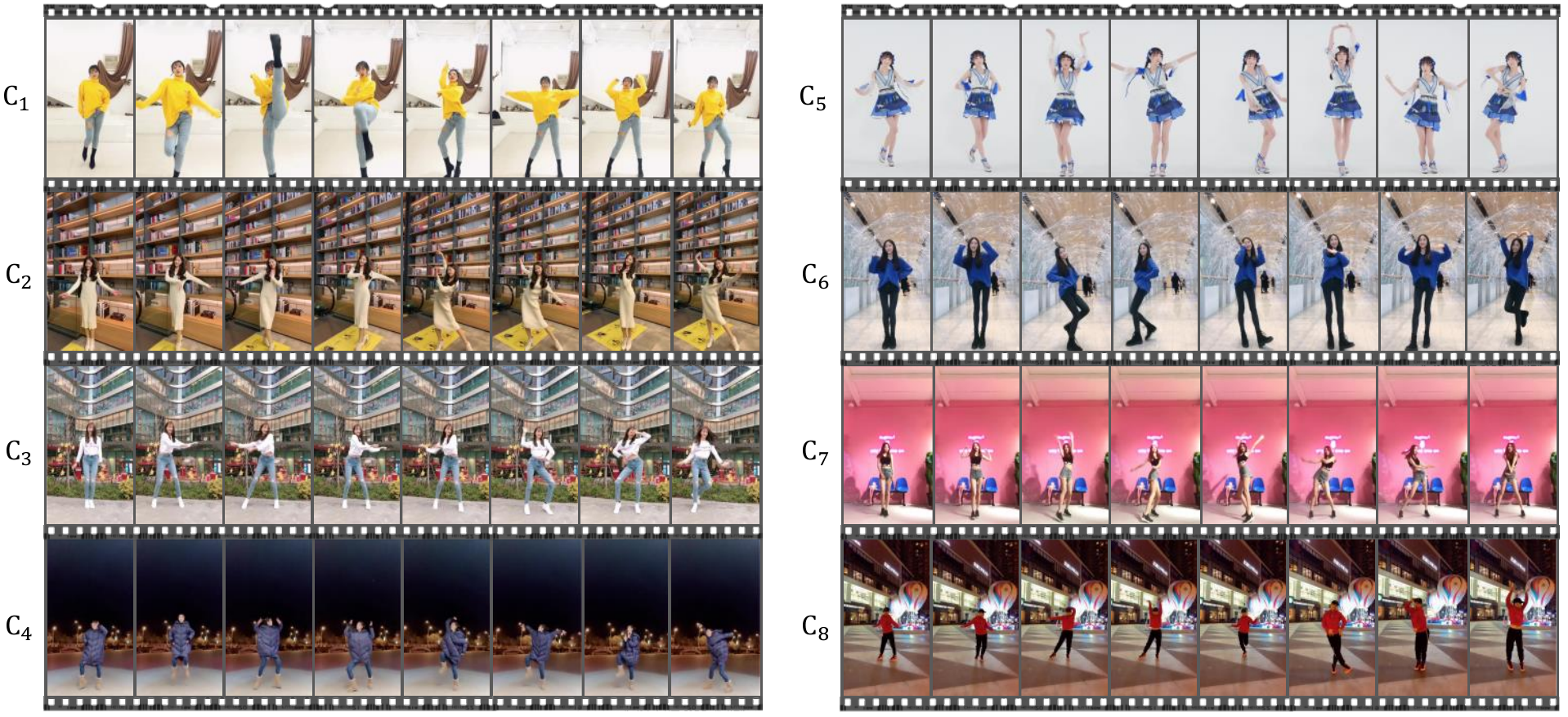}
\end{center}
   \vspace{-4mm}
   \caption{Example videos of dance challenges in VDV dataset. Eight dance challenges ($C_1{\sim}C_8$) with diverse dance styles are included. Please refer to our supplementary material for more details.}
   \vspace{-5mm}
\label{fig:vdv}
\end{figure*}

The performance scores in Olympic events~\cite{TP-toolbox-web} and surgical skills~\cite{martin1997objective} are given by the professionals. Actions of individual performers are usually carried out rigorously in similar environments. Only body movements are of major concern in these tasks.
In comparison, viral dance is a more improvisational kind of activity. Although the music is the same in each challenge, the expression can be very dissimilar among dancers. Additionally, every dancer has unique appearance of stature, clothing, backgrounds, etc. It hence requires analysis from multiple features for comprehensive virality prediction.
In this work, we introduce a multi-modal framework to capture virality information from both skeletal and appearance modalities. We demonstrate that the proposed PSGCN and RTCN models are more effective in dance virality prediction compared with state-of-the-art 3D-CNN based methods. Furthermore, the scale of our VDV dataset is much larger than existing human-centric performance assessment datasets~\cite{pirsiavash2014assessing,gao2014jhu,parmar2017learning,parmar2019action,doughty2019pros,gao2020asymmetric}, which makes it more challenging in terms of model robustness and generalization ability.


\section{Viral Dance Video Dataset}
\label{sec:data_collection}

\subsection{Video Collection}
\label{subsec:selection_criteria}

Our viral dance video (VDV) dataset is collected from the most popular short video app, TikTok. To begin with, we select eight representative viral dance challenges with exclusive background music. As presented in Fig.~\ref{fig:vdv}, these challenges have characteristic body movements and diverse dance styles. More importantly, as shown in Table~\ref{tab:dataset}, every challenge has millions of view counts, which makes them among the most popular challenges at the time. For a fair virality comparison solely by the video content, we filter the uploaders with extremely high popularity (i.e. with over 1,000 followers). Although video clips on TikTok are limited to 3-15 seconds, we set the minimum video length to 5 seconds to avoid uninformative videos. Over 24,000 videos are collected initially. The videos are then manually screened to meet the requirement of virality prediction. Since we define dance virality prediction as a multi-modal human-centric performance assessment problem, the screening criteria are formalized as follows: 

\begin{itemize}
\item Only one single dancer is allowed to appear in the video.
\item The dancer is performing the same dance of the hashtagged TikTok challenge with the same background music.
\item At least the joints above the waist of the dancer, i.e. head top, upper neck, shoulders, elbows, hips as defined in the MPII~\cite{andriluka20142d} human pose dataset, are visible in the video.
\item The camera motion shall not disrupt the continuity of dance performance.
\end{itemize}

\renewcommand\arraystretch{0.88}
\begin{table}
\small
\caption{General Statistics of VDV dataset.}
\vspace{-4mm}
\begin{center}
\setlength{\tabcolsep}{2mm}{
\begin{tabular}{ l  c  c  c  c  c  c  c  c  c}
\toprule
Dance Challenge & $C_1$ & $C_2$  & $C_3$  & $C_4$  & $C_5$  & $C_6$  & $C_7$ & $C_8$ & Total \\ \midrule
\#Videos & 1,447 & 932 & 735 & 467 & 239 & 230 & 128 & 114 & 4,292\\ 
View count (in millions) & 3,223 & 32 & 622 & 225 & 184 & 216 & 715 & 153 & 5,370\\ 
Video length (in hours) & 5.1 & 3.9 & 3.1 & 2.0 & 1.0 & 1.0 & 0.5 & 0.5 & 17.1\\
\#Frames (in thousands) & 542 & 349 & 275 & 175 & 89 & 86 & 48 & 42 & 1,606\\ 
\#Human skeletons (in thousands) & 487 & 314 & 248 & 157 & 80 & 77 & 43 & 38 & 1,444\\ 
\#Facial images (in thousands) & 52 & 42 & 33 & 21 & 11 & 10 & 6 & 5 & 180\\ 
\#Scenic images (in thousands) & 87 & 56 & 44 & 28 & 14 & 14 & 8 & 7 & 258\\ \bottomrule
\end{tabular}}
\vspace{-3mm}
\end{center}
\label{tab:dataset}
\end{table}

\renewcommand\arraystretch{0.88}
\begin{table}
\small
\caption{Comparison of different human-centric performance assessment datasets.}
\vspace{-4mm}
\begin{center}
\begin{tabular}{ l c  c  c  c  c }
\toprule
Dataset & Content & Type & Multi-Modal & \#Categories & \#Videos \\ \midrule
MIT-Dive~\cite{pirsiavash2014assessing} & Diving & Body & $\times$ & 1 & 159 \\
JIGSAWS~\cite{gao2014jhu} & Surgery & Hand-centric & \checkmark & 3 & 39  \\
UNLV-Sports~\cite{parmar2017learning} & Olympic events & Body & $\times$ & 3 & 717  \\ 
AQA-7~\cite{parmar2019action} & Olympic events & Body & $\times$ & 7 & 1,189 \\
BEST~\cite{doughty2019pros} & Daily activities & Hand-centric & $\times$ & 7 & 216 \\
TASD-2~\cite{gao2020asymmetric} & Diving & Body & $\times$ & 2 & 303 \\
VDV (ours) & Viral dance &	Body & \checkmark &8 & 4,292 \\
\bottomrule
\end{tabular}
\vspace{-4mm}
\end{center}
\label{tab:datasets}
\end{table}

\subsection{Virality Score}
\label{subsec:virality_score}

While gathering videos, associated view counts are collected as well. We adopt a similar computation method of the ground-truth virality score as in~\cite{chen2016micro}. Our virality score is based on three popularity-related indicators, namely, like counts, comment counts and repost counts. We denote these indicators as $n_{like}$, $n_{comment}$ and $n_{repost}$. It is observed that the popularity of a short video accumulates
rapidly on a daily basis in a period of time (usually about one week)~\cite{li2018studies}. So we normalize the virality score with clipped existing time (in days), which is denoted as $d_{upload}$. A log transform is finally applied as
in~\cite{bielski2018pay,bielski2018understanding} to handle the large variation and skewness of the original score distribution. We define the raw virality score $s_r$ as:
\begin{equation}
s_r = log_{2}({\frac{n_{like}+n_{comment}+n_{repost}+1}{clip(d_{upload},1,7)}}),
\label{eqn1}
\end{equation}
where $clip$ denotes the clipping operation with min-max thresholds. The additional increment in the numerator prevents computing logarithm of zero. Since different dance challenge has inherent difference in popularity, for an accurate virality comparison across all challenges, we further normalize $s_r$ with min-max normalization to get the final virality score $s$, which is formulated as:
\begin{equation}
    s = \frac{s_{r}-min(s_{r})}{max(s_{r})-min(s_{r})}.
\label{eqn1_1}
\end{equation}
We denote the minimum and maximum raw score values of each challenge as $min(s_{r})$ and $max(s_{r})$. This min-max normalization makes the final virality score a value between 0 and 1, which is compatible for both intra- and inter-challenge virality comparison. 

\subsection{Dataset Overview and Comparison}

We present general statistics of VDV dataset in Table~\ref{tab:dataset}. VDV contains 4,292 viral dance clips of eight challenges ($C_1{\sim}C_8$). The total length of video data is $17.1$ hours with over $1.6$ million video frames. Moreover, as later introduced in Section~\ref{subsec:appearance-based prediction}, we utilize state-of-the-art deep learning models~\cite{fang2017rmpe,wang2016temporal,cao2018vggface2,zhou2017places} to extract multi-modal (i.e. skeletal, holistic, facial and scenic) features from RGB videos and release them as a part of our dataset. The numbers of extracted frames and skeletal, facial, scenic features in each challenge are listed in Table~\ref{tab:dataset}. 
With abundant ready-to-use features in each modality, 
multi-modal solutions can be adequately evaluated on our VDV benchmark.

Table~\ref{tab:datasets} shows the comparison of VDV dataset and other publicly available human-centric performance assessment datasets. These datasets are from various domains including sports~\cite{pirsiavash2014assessing, parmar2017learning, parmar2019action,gao2020asymmetric}, surgical~\cite{gao2014jhu} and daily activities~\cite{doughty2019pros}. However, online short videos with great commercial value like viral dance challenges have not been explored before VDV dataset. Datasets like JIGSAWS~\cite{gao2014jhu} and BEST~\cite{doughty2019pros} are hand-centric rather than focusing on the entire human body, which limits their applicability in common scenarios. In terms of data modality, JIGSAWS dataset provides kinematic and video data captured by a surgical system, while other datasets only provide video data except VDV. In terms of dataset scale, to our best knowledge, VDV is currently the largest human-centric performance assessment benchmark.

\section{Approach}
\label{sec:approach}

\begin{figure*}
\begin{center}
   \includegraphics[width=0.7\textwidth]{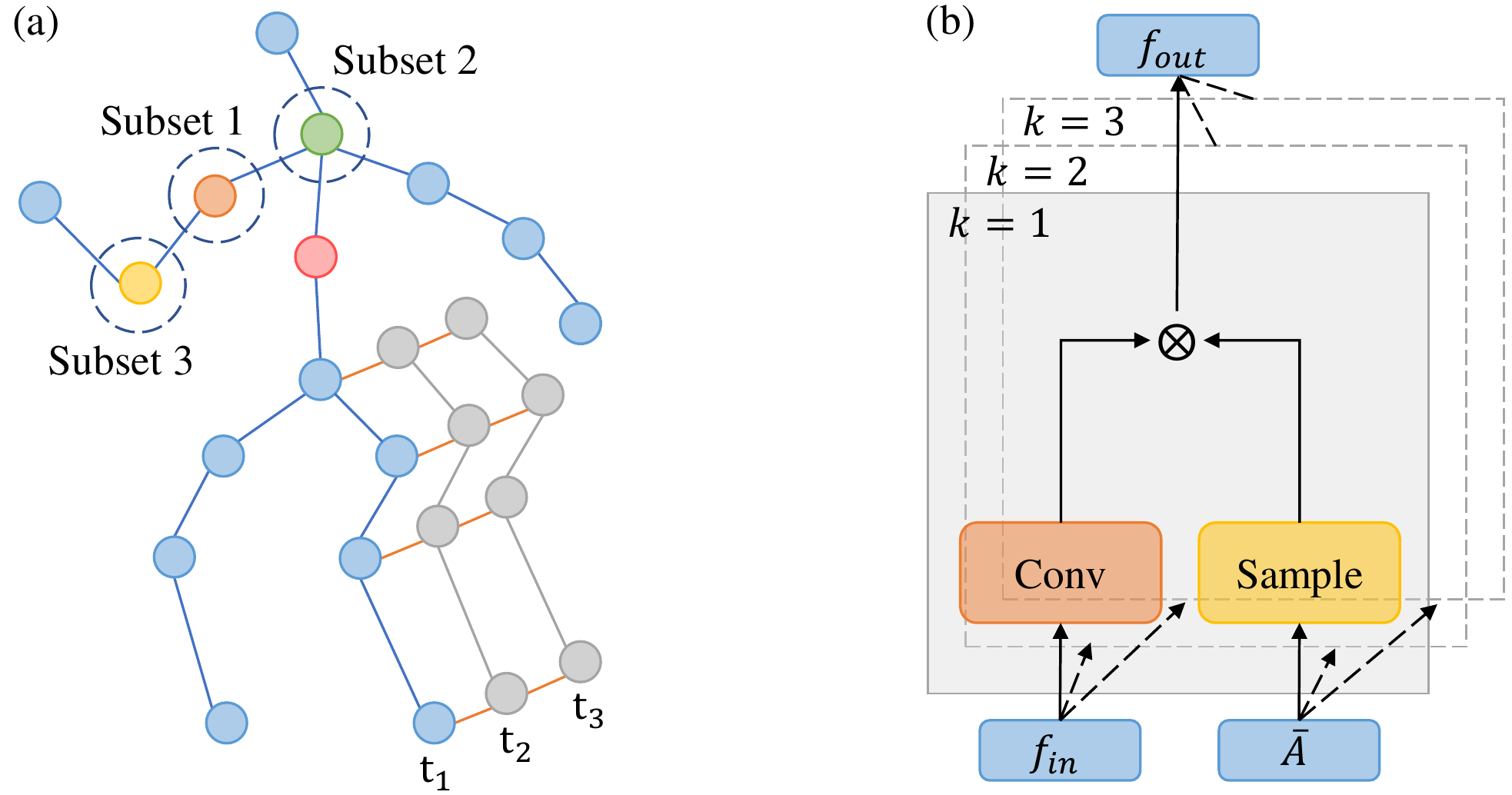}
\end{center}
\vspace{-3mm}
   \caption{(a) Spatial-temporal skeleton graph. Human joints in the same frame are connected with spatial edges (blue). Each joint is connected to itself in consecutive frames with temporal edges (orange). Only left leg joints in three consecutive frames ($t_1\sim{t_{3}}$) are shown in this figure. The neighboring area of each vertex is partitioned into three subsets determined by the distance to center joint of thorax (red). We show neighboring subsets of the right shoulder joint (orange) in this figure. (b) Graph down-sampling module. The normalized adjacency martix $\overline{A}$ is refined with $sample$ operations performed parallelly on $K$ subsets.}
   \vspace{-4mm}
\label{fig:gcn}
\end{figure*}

\subsection{Skeleton-based Prediction}
\label{subsec:Skeleton-based_prediction}

To explore human body movements, we directly utilize skeleton features, i.e. the 2D coordinates of body joints estimated by the state-of-the-art pose estimator AlphaPose~\cite{fang2017rmpe}. As human skeleton sequence is naturally non-Euclidean, we employ graph
convolutions~\cite{kipf2016semi,niepert2016learning,yan2018spatial,li2019actional} and propose a pyramidal skeleton graph convolutional network (PSGCN) for skeleton-based virality prediction.

\noindent\textbf{Spatial-temporal graph convolution.} Yan et al.~\cite{yan2018spatial} first propose to use a spatial temporal graph convolutional network (ST-GCN) to model skeleton data for action recognition problem. As shown in Fig.~\ref{fig:gcn} (a), we adopt a similar graph structure as their work, where joints in the same frame are spatially connected  with spatial edges following the definition in MPII dataset~\cite{andriluka20142d}. For the temporal dimension, the same joints in two consecutive frames are connected with temporal edges. The skeleton features are then expressed as a tensor $f\in{{\mathbb{R}}^{T\times{V}\times{C}}}$, where $T$, $V$, $C$ denote temporal length, number of vertexes and number of channels, respectively. We define the graph convolution on such a spatial-temporal graph as:
\begin{equation}
    f_{out}(v_{ti})=\sum_{v_{tj}\in{B(v_{ti})}}{\frac{1}{Z_{ti}(v_{tj})}f_{in}(v_{tj})\cdot{w(l_{ti}(v_{tj}))}},
\label{eqn2}
\end{equation}
where $v_{ti}$ is the vertex of the $i$-th joint in frame $t$ and $l$ is the graph labeling function~\cite{niepert2016learning} that helps to gather corresponding weights from the convolution kernel $w$. $B$ denotes the neighboring area of each vertex. As Fig.~\ref{fig:gcn} (a) shows, $B$ is partitioned into $K=3$ subsets following the spatial partitioning strategy defined in~\cite{yan2018spatial}, which divides neighboring joints according to their distance to the center joint. The cardinality of each subset, denoted as $Z$, is employed as a normalization term. We adopt the same implementation of graph convolution as in~\cite{kipf2016semi}. For the spatial dimension, the graph convolution is computed as:
\begin{equation}
    f_{out}=\sum_{k=1}^{K}{\Lambda_{k}^{-\frac{1}{2}}A_{k}\Lambda_{k}^{-\frac{1}{2}}f_{in}\ast{w_{k}}},
\label{eqn3}
\end{equation}
where $A_{k}\in{{\mathbb{R}}^{V\times{V}}}$ is the adjacency matrix of the $k$-th subset. $\Lambda_{k}^{ii}=\sum_{j}(A_{k}^{ij})$ is utilized to normalize the adjacency matrix. $w_{k}\in{{\mathbb{R}}^{C_{in}\times{C_{out}}\times{1}\times{1}}}$ is the convolution kernel. The input feature map $f_{in}\in{{\mathbb{R}}^{T\times{V}\times{C_{in}}}}$ is convolved with $w_k$ by the convolution operation $\ast$. For the temporal dimension, we directly perform a $K_t\times{1}$ convolution on the output feature map $f_{out}\in{{\mathbb{R}}^{T\times{V}\times{C_{out}}}}$ with temporal kernel size $K_t$, because only the same joints in two consecutive frames are connected.

\begin{figure*}
\begin{center}
   \includegraphics[width=0.9\textwidth]{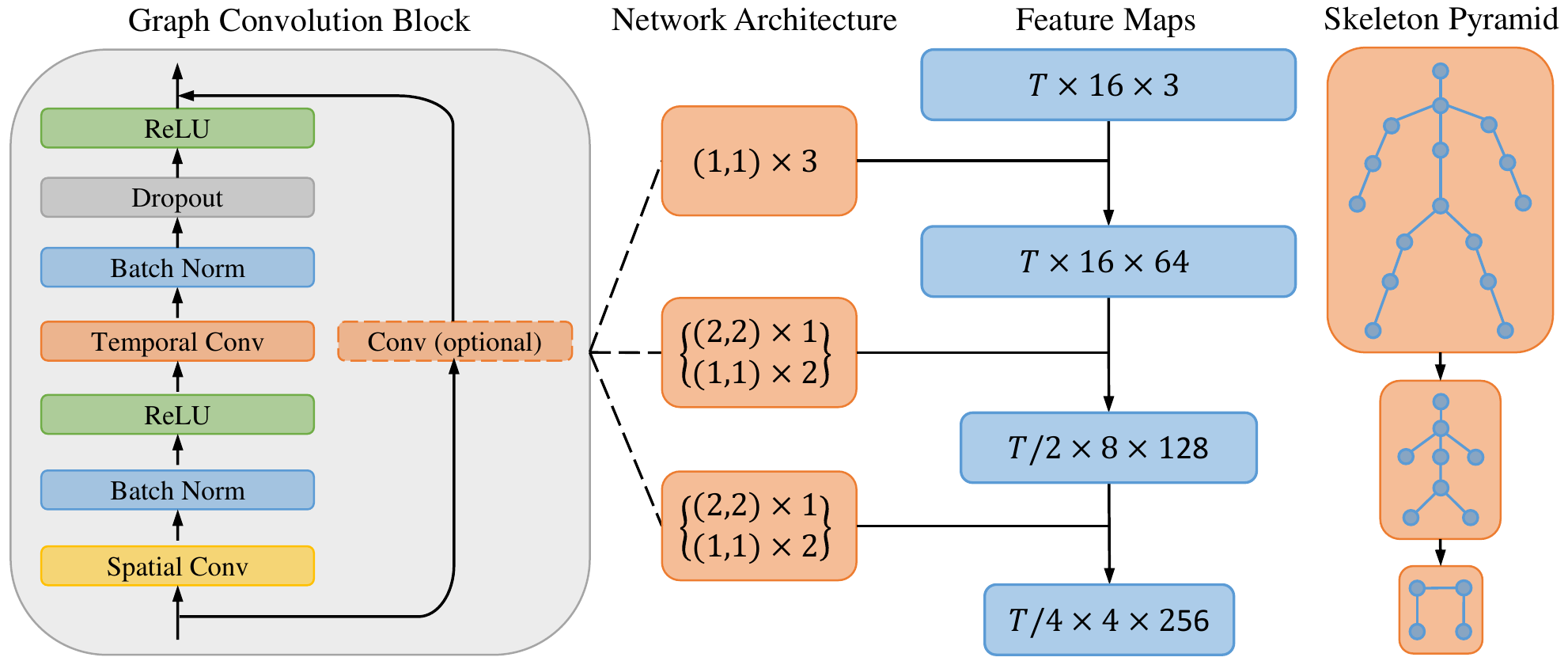}
\end{center}
   \vspace{-3mm}
   \caption{Illustrations of graph convolution block, network architecture, feature maps and skeleton pyramid in PSGCN (from left to right). PSGCN is comprised of 9 graph convolution blocks, whose parameters are shown as \emph{(temporal stride, spatial stride) $\times$ block number}. The sizes of graph feature maps are given as \emph{temporal length $\times$ number of vertexes $\times$ number of channels}.}
   \vspace{-5mm}
\label{fig:gcn2}
\end{figure*}

\noindent\textbf{Pyramidal skeleton graph convolutional network.} With the spatial-temporal graph convolution defined above, we devise the pyramidal skeleton graph convolutional network (PSGCN) which explicitly refines skeleton graphs to capture virality information. It is demonstrated that reducing the resolution of feature maps with pooling layers improves the performance and robustness of CNN models~\cite{krizhevsky2012imagenet,yu2015multi}. However, it is nontrivial to extend the pooling operation to graph convolutional networks as the graph nodes have no locality information. As shown in Fig.~\ref{fig:gcn} (b), we propose a graph down-sampling module for skeleton graphs to address this problem. Let $f_{in}\in{{\mathbb{R}}^{T\times{V}\times{C_{in}}}}$ and $\overline{A}\in{{\mathbb{R}}^{K\times{V}\times{V}}}$ denote the input feature map and normalized adjacency matrix, respectively. The graph down-sampling operation is performed parallelly on $K$ subsets as:
\begin{equation}
    f_{out}=\sum_{k=1}^{K}{sample({\overline{A}}_{k})f_{in}\ast{w_{k}^{d}}},
\label{eqn3_2}
\end{equation}

\begin{equation}
    sample(A)^{i,j}=\frac{1}{4}
    {\begin{bmatrix} i+2 & -i \end{bmatrix}}
    {\begin{bmatrix} 
        A^{2i,2j} & A^{2i, 2(j+1)}\\
        A^{2(i+1),2j} & A^{2(i+1), 2(j+1)}
     \end{bmatrix}}
    {\begin{bmatrix} j+2 \\ -j \end{bmatrix}},
    \label{eqn_sample}
\end{equation}
where we refine $\overline{A}$ with the $sample$ operation and employ a learnable convolution kernel $w_{k}^{d}\in{{\mathbb{R}}^{C_{in}\times{C_{out}\times{2}\times{1}}}}$ with a stride of two. As formulated in Eqn.~\eqref{eqn_sample}, the $sample$ operation produces down-sampled adjacency matrices by quadratically interpolating between neighboring vertexes.
Finally, outputs of $K$ subsets are fused by element-wise summation, resulting in the output feature map $f_{out}\in{{\mathbb{R}}^{T\times{\frac{V}{2}}\times{C_{out}}}}$.

Our PSGCN is composed of 9 graph convolution blocks as illustrated in Fig.~\ref{fig:gcn2}, where parameters of the network architecture are given as \emph{(temporal stride, spatial stride)$\times$block number}. Within each convolution block, we employ the residual learning mechanism~\cite{he2016deep}, in which an optional $1\times1$ convolutional layer is utilized when input and output channels are not identical. To increase the flexibility of our model, we introduce learnable edge weighting into PSGCN. The adjacency matrix in Eqn.~\eqref{eqn3_2} is actually formed as $\overline{A} = \overline{A}_{o} + A_{l}$, where $\overline{A}_{o}$ is the original normalized adjacency matrix and $A_{l}$ is the learnable connectivity matrix parameterized for each graph convolution block. With this learnable weighting mechanism, our model can partially adjust the graph structure according to the semantic information in different graph convolution blocks. 
As Fig.~\ref{fig:gcn2} shows, the skeleton graphs in PSGCN are hierarchically refined, forming a pyramidal network architecture.
We demonstrate the advantage of this pyramidal skeleton architecture later in Section~\ref{subsec:virality_prediction_results}.

\subsection{Action Feedback} 
\label{subsec:action_feedback_intro}

Besides virality prediction, our PSGCN is also capable of generating action feedback for dancers on how to improve their performance. Although teaching videos of viral dance challenges are uploaded by professional dancers in short video communities sometimes, 
making a high quality teaching video is usually laborious and requires extensive experience.
Also, it is difficult for starters to obtain personalized advice on body movements in these videos. Therefore, automatic action feedback for dance challenges is a very promising application. In this section, we present the generation of two kinds of practical action feedback, i.e. action guidance and dance tips. 

\noindent\textbf{Action guidance.} As PSGCN takes human joint coordinates as input to predict the virality score, action guidance leading to higher score can be acquired by differentiating the predicted virality score w.r.t. the joint coordinates. Concretely, let $s$ denote the predicted score for a dance clip and $v_{tj}$ denote the graph vertex containing 2D coordinate of the $j$-th joint in frame $t$, we derive the gradient of predicted score w.r.t. each joint position as:
\begin{equation}
    g_{tj} = [x_{tj}, y_{tj}] = \frac{\partial{s}}{\partial{v_{tj}}}.
    \label{eqn_feedback_1}
\end{equation}
We define $g_{tj}$ as the action guidance vector of the $j$-th joint in frame $t$. These vectors indicate the direction toward which the dancers shall adjust their body to improve the score. Compared with the man-made teaching videos, our action guidance is generated on-the-fly and personalized for individual users.

\noindent\textbf{Dance tips.} It is beneficial for most challengers to be aware of dance tips on crucial body parts before performing the dance. 
Here we derive two kinds of dance tips from our model, i.e. temporal attention and overall attention. 
The first one shows the temporal variation of body-part importance in each challenge while the second one indicates the overall body-part influence across challenges.
Let $g_{itj}$ denote the action guidance vector (defined in Eqn.~\eqref{eqn_feedback_1}) of the $i$-th video in challenge $c$, we compute the temporal attention as:
\begin{equation}
    m_{ctj} = \sum_{i=1}^{N}\left\|{g_{itj}}\right\|_2/N,
    \label{eqn_feedback_2}
\end{equation}
where $N$ is the video number of challenge $c$. Consequently, $m_{ctj}$ is mathematically the averaged feedback magnitude of the $j$-th joint at frame $t$ in challenge $c$, where the magnitude is represented by the $\ell_2$ norm of feedback vectors.
Temporal attention provides dancers with specific tips by telling them which body part should be paid more attention at different moment.
Let $T_i$ denote the frame number of the $i$-th video in $c$, the overall attention is computed as:
\begin{equation}
    m_{cj} = \sum_{i=1}^{N}(\sum_{t=1}^{T_i}\left\|{g_{itj}}\right\|_2/{T_{i}})/N.
    \label{eqn_feedback_3}
\end{equation}
Overall attention gives dancers general tips on which body part deserves more attention when participating in different challenges. Examples of above feedback results are presented in Section~\ref{subsec:action_feedback}.

\subsection{Appearance-based Prediction}
\label{subsec:appearance-based prediction}

In terms of dance virality prediction, visual appearance is also a strong reference and usually complementary to skeleton features. Therefore, we incorporate the appearance-based prediction streams in our framework. As shown in Fig.~\ref{fig:framework}, three streams of appearance features, i.e. holistic, facial and scenic streams, are utilized for comprehensive appearance-based virality prediction.

\noindent\textbf{Holistic features.} Video frames extracted directly from dance clips contain the holistic appearance information of the whole picture. We use a TSN~\cite{wang2016temporal} model pre-trained
on the Kinetics~\cite{carreira2017quo} dataset to extract holistic appearance features from video frames at 25-fps. Each frame is represented by a 2,048 dimensional vector.

\noindent\textbf{Facial features.} As shown in Fig.~\ref{fig:intro}, dance clips tend to be more attractive when the dancer has a gorgeous face or attractive expressions. Therefore, we take facial features into
consideration in virality prediction. We use MTCNN~\cite{zhang2016joint} to detect human face at 3-fps for each video and then extract facial features with a ResNet-50~\cite{he2016deep} model pre-trained on VGGFace2~\cite{cao2018vggface2} dataset. Finally, every face image is represented by a 2,048 dimensional vector.

\noindent\textbf{Scenic features.} We also observe that a good background scenery helps to increase the virality of videos. To capture scenic features, we first perform background extraction with the ViBe algorithm~\cite{van2012background} performed in a sliding window of 120 frames. After that, a ResNet-18~\cite{he2016deep} model pre-trained on the scene recognition dataset Places~\cite{zhou2017places} is used to extract the scenic features. In the end, we obtain a 512 dimensional feature vector for each background image.

\begin{figure*}
\begin{center}
   \includegraphics[width=1.0\textwidth]{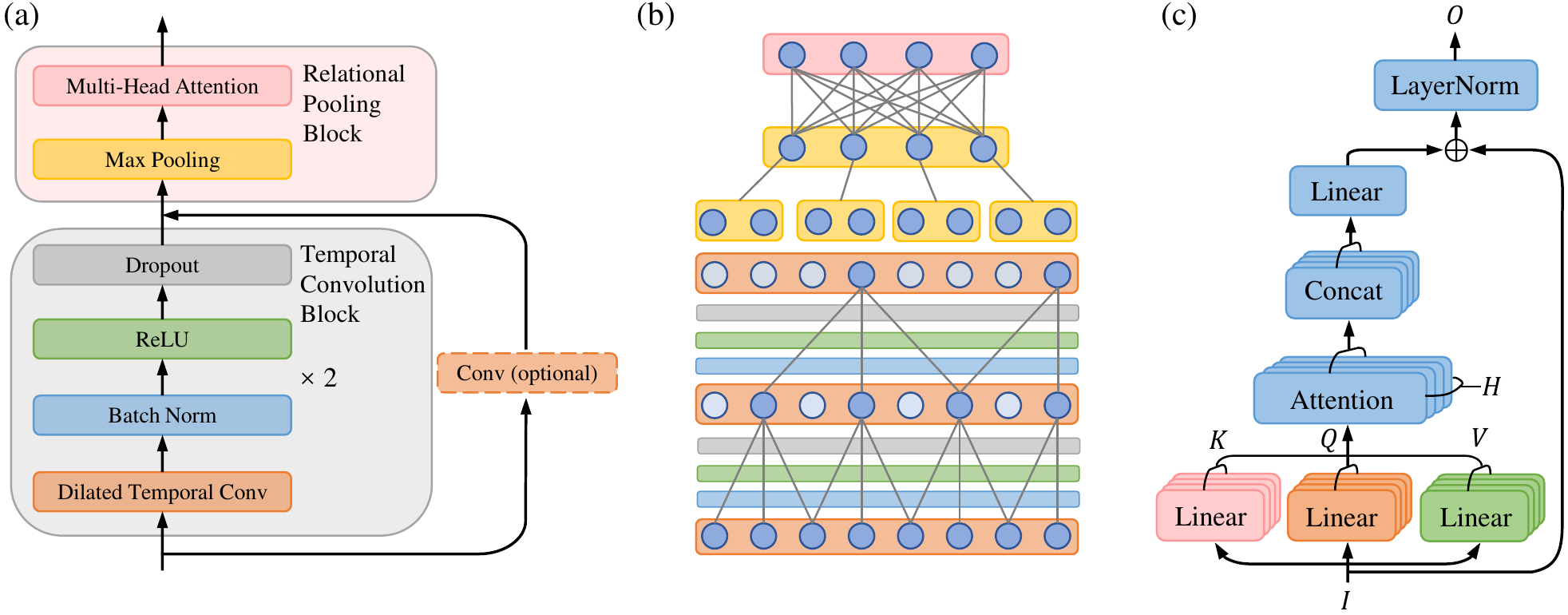}
\end{center}
   \vspace{-4mm}
   \caption{Relational temporal convolutional network. (a) Structures of temporal convolution block and relational pooling block. (b) Simplified illustration of neural connections in RTCN. Note that in the relational pooling block, each temporal position is adaptively connected with all position from the previous layer, facilitated by multi-head self-attention. (c) Multi-head self-attention. Scaled dot-product attentions are computed parallelly on $H$ attention heads.}
   \vspace{-6mm}
\label{fig:tcn}
\end{figure*}

\noindent\textbf{Relational temporal convolutional network.} As the appearance features are extracted at frame-level, it is crucial to capture temporal dynamics from these features. For example, the variations in facial expressions can only be exploited by modeling a sequence of consecutive facial images. Recent studies have demonstrated that temporal convolutional networks (TCN) are of great competitiveness in plenty of sequence modeling tasks including action segmentation~\cite{lea2017temporal,farha2019ms,wang2019atrous}, natural language processing~\cite{dauphin2017language,bai2018empirical} and so on. Compared with recurrent neural network (RNN) architectures, TCN has lower memory requirement and less training difficulty~\cite{bai2018empirical}. 
However, conventional TCN models are subject to the locality of convolution operations, making them weak at perceiving non-local frame-to-frame relations in video understanding tasks. 
We believe that such non-local relations are important in dance virality prediction as well. For example, in some dance challenges, there exist corresponding relations between the starting and ending actions of the dancer. These distant relations is hard to be captured with conventional TCNs.
Therefore, we propose a relational temporal convolutional network (RTCN), 
which excels in capturing non-local relations between distant temporal positions while retaining the efficiency of TCN architectures.

As illustrated in Fig.~\ref{fig:tcn} (a), the basic components of our RTCN are temporal convolution blocks and relational pooling blocks. The temporal convolution block is designed under the paradigm of residual convolution blocks~\cite{he2016deep}, which has two consecutive sequences of dilated temporal convolution, batch normalization, ReLU and dropout layers. The input of each block is processed by an optional $1\times1$ convolutional layer and added to the output. Fig.~\ref{fig:tcn} (b) shows a simplified example of neural connections within the structure of Fig.~\ref{fig:tcn} (a). The convolution kernel size is 3 and the dilation rates are exponentially increased. 
In practice, we set the dilation rates to $2^n$, where $n$ is the number of the corresponding convolution block.

To explicitly capture non-local relations upon temporal convolution features, we devise the relational pooling block as shown in Fig.~\ref{fig:tcn} (a).
The relational pooling block is composed of a max pooling layer and a multi-head self-attention layer. 
The max pooling layer improves feature robustness by reducing the temporal length. 
These features are then fed into a multi-head self-attention layer, where holistic temporal relations are explored. 
As presented in Fig.~\ref{fig:tcn} (b), each temporal position is adaptively connected to all positions from previous features under the self-attention mechanism, which is introduced later.
In summary, our RTCN is composed of six convolution blocks with channel numbers of $\{256, 256, 512, 512, 1024, 1024\}$. We insert three relational pooling blocks after the second, forth and sixth convolution blocks, respectively.

\noindent\textbf{Multi-head self-attention.} Self-attention mechanism~\cite{vaswani2017attention} is demonstrated to be a powerful tool for non-local feature aggregation in many research fields including machine translation~\cite{vaswani2017attention,devlin2018bert}, action recognition~\cite{wang2018non} and object detection~\cite{carion2020end}. In this work, we introduce a multi-head self-attention layer in the relational pooling block to adaptively extract non-local features. 
As is shown in Fig.~\ref{fig:tcn} (c), let $I\in{{\mathbb{R}}^{T\times{C}}}$ denote the $input$ tensor of the self-attention layer, it is first parallelly projected into $H$ sets of $query$, $key$, $value$ tensors. We denote each of these triplets as $Q, K, V\in{{\mathbb{R}}^{T\times{C^{'}}}}$, where ${C^{'}} = C/H$. The non-local features are computed with scaled dot-product attention~\cite{vaswani2017attention} as:
\begin{equation}
    Attention(Q, K, V) = softmax({\frac{Q{K^{T}}}{\sqrt{C^{'}}}})V.
\label{eqn4}
\end{equation}
Here the compatibility of $Q$ and $K$ is computed by matrix dot product and scaled by ${\sqrt{C^{'}}}$ to avoid exploding gradients. The $softmax$ function is used to obtain normalized attention weights, which are used to compute a weighted sum of $V$. Outputs of $H$ attention heads are then concatenated and once again projected. After a residual connection followed by layer normalization~\cite{ba2016layer}, we obtain the final output $O\in{{\mathbb{R}}^{T\times{C}}}$. This multi-head self-attention process is formulated as:
\begin{equation}
    O = {LayerNorm}({Concat}_{i=1}^{H}({Attention}(I{W}_{i}^{Q}, I{W}_{i}^{k}, I{W}_{i}^{V})){W^{O}}+I),
\label{eqn5}
\end{equation}
where $W^{Q}, W^{K}, W^{V}\in{{\mathbb{R}}^{H\times{C}\times{C^{'}}}}$ are the multi-head linear projection matrices and $W^{O}\in{{\mathbb{R}}^{{C}\times{C}}}$ is the output projection matrix. The concatenation of $H$ tensors are denoted as ${Concat}_{i=1}^{H}$. We utilize $H=4$ attention heads and visualize the results in Section~\ref{subsec:virality_prediction_results}.

\subsection{Attentive Modality Fusion}
\label{subsec:attentive_modality_aggregation}

A dance clip can go viral for a number of reasons. For example, some videos are popular for exaggerations in body movements while others are popular for the attractive appearance of dancers.
Furthermore, in terms of different dance challenge, the importance of different visual aspect varies as well. 
As Fig.~\ref{fig:framework} shows, we adopt a late fusion (score fusion) strategy facilitated by a modality attention mechanism.
There are two major advantages of adopting this late fusion strategy. Firstly, it avoids feature degradation in feature-level fusion schemes, as dimensionalities of multi-modal features have significant difference (up to 8x as shown in Fig~\ref{fig:framework}). Secondly, it ensures each modality stream makes prediction independently, facilitating practical applications like multi-dimensional video recommendation. 
Concretely, we concatenate video-level features of different modalities and feed it into the modality attention module, which consists of two fully-connected (FC) layers followed by a $softmax$ function. The modality attention weight vector $A$ is derived as:
\begin{equation}
    A = [a_{p}, a_{h}, a_{f}, a_{b}] = softmax(I{W}^{H}{W}^{O}),
    \label{eqn6}
\end{equation}
where $a_p$, $a_h$, $a_f$, $a_b$ denote attention weights for the skeletal, holistic, facial and scenic modalities, respectively. $I\in{{\mathbb{R}}^{1\times{C}}}$ is the input $C$ dimensional concatenated feature vector. $W^{H}\in{{\mathbb{R}}^{C\times{C^{'}}}}$ and $W^{O}\in{{\mathbb{R}}^{{C^{'}}\times{4}}}$ are the parameter matrices of the hidden and output FC layers. Let $S = [s_p, s_h, s_f, s_b]$ denote the predicted scores from corresponding modality streams in $A$, the final virality score prediction $s$ is computed as: $s=AS^{T}$.
To stabilize the training of different modality streams, we impose ranking constrains on each of them as later introduced in Section~\ref{subsec:implementation_and_experimental_setup}.

\section{Experiments}
\label{sec:experiments}

\renewcommand\arraystretch{0.9}
\begin{table*}
\small
\caption{Comparison results on VDV dataset. Results of the skeleton-based, appearance-based (holistic and multi-modal features) and integrated experiments evaluated by SRC are listed from top to bottom. For the last two experiments, black, red and green numbers indicate the performance is \emph{retained}, \emph{\textcolor{red}{increased}} and \emph{\textcolor{green}{decreased}}, respectively. The best values in each experiment are highlighted in bold. Best viewed in color.}
\vspace{-3mm}
\begin{center}
\setlength{\tabcolsep}{0.58mm}{
\begin{tabu}{ l | c c c c | c c  c  c  c  c  c  c | c}
\tabucline[1.25pt]{-}
\multirow{2}{*}{Method} & \multicolumn{4}{c|}{Feature Modalities} & \multicolumn{8}{c|}{Single-Challenge} & All- \\
 & Skeletal & Holistic & Facial & Scenic & $C_1$ & $C_2$  & $C_3$  & $C_4$  & $C_5$  & $C_6$  & $C_7$ & $C_8$ & Challenge\\
\hline
Pose+DCT~\cite{pirsiavash2014assessing} & \checkmark &  &  &  & -0.03 & 0.02 & 0.06 & -0.02 & 0.28 & -0.07 & 0.35 & 0.22 & 0.04\\
ST-GCN~\cite{yan2018spatial} & \checkmark &  &  &  & 0.23 & 0.14 & 0.21 & 0.39 & 0.31 & 0.27 & 0.49 & 0.38 & 0.24\\
DMGNN~\cite{li2020dynamic} & \checkmark &  &  &  & \textbf{0.25} & \textbf{0.17} & \textbf{0.23} & 0.38 & 0.31 & 0.29 & 0.45 & 0.36 & 0.25\\
PSGCN (ours) & \checkmark &  &  &  & \textbf{0.25} & 0.15 & 0.22 & \textbf{0.42} & \textbf{0.33} & \textbf{0.30} & \textbf{0.51} & \textbf{0.41} & \textbf{0.26}\\ 
\hline
C3D+SVR~\cite{parmar2017learning} &  & \checkmark &  &  & 0.10 & 0.06 & 0.25 & 0.37 & -0.09 & 0.28 & 0.41 & 0.15 & 0.16\\
I3D+SVR~\cite{carreira2017quo} &  & \checkmark &  &  & 0.12 & 0.09 & 0.23 & 0.36 & 0.05 & 0.29 & 0.39 & 0.19 & 0.18\\
SlowFast+SVR~\cite{feichtenhofer2019slowfast} &  & \checkmark &  &  & 0.15 & 0.11 & 0.25 & 0.36 & 0.10 & 0.31 & 0.40 & 0.23 & 0.21\\
S3D~\cite{xiang2018s3d} &  & \checkmark &  &  & 0.14 & 0.07 & 0.12 & 0.31 & 0.11 & 0.18 & \textbf{0.43} & 0.40 & 0.22\\ 
Spatial Attn.~\cite{bielski2018pay} &  & \checkmark &  &  & 0.18 & 0.09 & 0.19 & 0.17 & 0.14 & 0.26 & 0.38 & 0.35 & 0.20\\
USDL~\cite{tang2020uncertainty} &  & \checkmark &  &  & 0.27 & 0.14 & \textbf{0.27} & 0.33 & 0.17 & 0.31 & 0.39 & 0.41 & 0.24 \\
LSTM~\cite{hochreiter1997long} &  & \checkmark &  &  & 0.12 & 0.07 & 0.16 & 0.24 & 0.10 & 0.23 & 0.37 & 0.39 & 0.17\\
TCN~\cite{bai2018empirical} &  & \checkmark &  &  & 0.25 & 0.13 & 0.22 & 0.35 & 0.18 & 0.30 & 0.41 & 0.42 & 0.25\\
RTCN (ours)&      & \checkmark &            &            & \textbf{0.29} & \textbf{0.17} & 0.24 & \textbf{0.39} & \textbf{0.20} & \textbf{0.33} & 0.42 & \textbf{0.44} & \textbf{0.27}\\
\hline
RTCN (ours)       &            & \checkmark & \checkmark &            & 0.29 & \textcolor{red}{0.23} & \textcolor{red}{0.26} & \textcolor{red}{0.41} & \textcolor{red}{\textbf{0.28}} & \textcolor{red}{0.35} & \textcolor{red}{\textbf{0.52}} & \textcolor{red}{0.49} & \textcolor{red}{0.30}\\
RTCN (ours)       &            & \checkmark &            & \checkmark & \textcolor{red}{0.32} & \textcolor{red}{0.21} & \textcolor{green}{0.21} & \textcolor{red}{0.42} & \textcolor{green}{0.19} & 0.33 & \textcolor{red}{0.46} & \textcolor{red}{\textbf{0.50}} & \textcolor{red}{0.28}\\
RTCN (ours)       &            & \checkmark & \checkmark & \checkmark & \textcolor{red}{\textbf{0.35}} & \textcolor{red}{\textbf{0.24}} & \textcolor{red}{\textbf{0.27}} & \textcolor{red}{\textbf{0.45}} & \textcolor{red}{0.27} & \textcolor{red}{\textbf{0.46}} & \textcolor{red}{0.47} & \textcolor{red}{0.49} & \textcolor{red}{\textbf{0.31}}\\ \hline
Integrated (ours) & \checkmark & \checkmark & \checkmark & \checkmark & \textcolor{red}{\textbf{0.37}} & 0.24 & \textcolor{red}{\textbf{0.30}} & \textcolor{red}{\textbf{0.47}} & 0.33 & \textcolor{red}{\textbf{0.48}} & \textcolor{red}{\textbf{0.54}} & \textcolor{red}{\textbf{0.53}} & \textcolor{red}{\textbf{0.34}}\\ 
\tabucline[1.25pt]{-}
\end{tabu}}
\vspace{-5mm}
\end{center}
\label{tab:experiment_result}
\end{table*}

\subsection{Experimental Setup}
\label{subsec:implementation_and_experimental_setup}

\noindent\textbf{Evaluation protocol.} To comprehensively evaluate the performance and generalization ability of different models, we set up single-challenge and all-challenge tracks of the VDV benchmark. In the single-challenge tracks, we separately split training and test sets for each challenge. Models are trained and tested on eight challenges respectively, evaluating the cross-challenge robustness of models and their performance with limited single-challenge data. In the all-challenge track, videos from eight challenges are shuffled and split altogether in order to evaluate models on the complete dataset. This track is more challenging as it evaluates the generalization ability of models and their performance under class-agnostic scenarios. The dataset partitions of all tracks are fixed with $80\%$ training data and $20\%$ testing data. As in many other score prediction tasks~\cite{bielski2018pay,pirsiavash2014assessing,parmar2017learning}, we use Spearman's rank correlation coefficient (SRC) as the performance metric, where a value of 1 represents the identical ranking order as the ground-truth and -1 represents the opposite.

\noindent\textbf{Loss functions.} Our loss function $L$ consists of a mean square error (MSE) term $L_s$ and margin ranking loss~\cite{burges2005learning,doughty2019pros} terms $L_r$, which are formulated as:
\begin{equation}
    L_{r} = \sum_{i=1}^{N}\sum_{j=i+1}^{N}max((p_i-p_j)sign(g_j-g_i)+\delta,0),
    \label{eqn7}
\end{equation}
and
\begin{equation}
    L = \alpha{L_{s}}+\beta{L_{r}}+\gamma{\frac{1}{M}\sum_{m=1}^M{L_{r}^m}},
    \label{eqn8}
\end{equation}
where $p_i$ and $g_i$ denote the predicted and ground-truth virality scores for sample $i$ in a training batch of size $N$. $sign$ is the sign taking operation and $\delta$ denotes the constant margin. $L_s$ and $L_r$ are applied on the final prediction $s$ as introduced in Section~\ref{subsec:attentive_modality_aggregation}. Additionally, we add a ranking loss term $L_{r}^m$ to each of the $M$ modality streams to encourage right ranking order without constraining the score values. $\alpha$, $\beta$ and $\gamma$ stand for the weights of corresponding loss terms, which are automatically adjusted with the Gradnorm~\cite{chen2018gradnorm} algorithm.

\noindent\textbf{Implementation and competitor methods.} Our framework is implemented with PyTorch~\cite{paszke2017automatic}. The temporal kernel size is set to 7 for both PSGCN and RTCN models. $1\times1$ convolution layers are added at the beginning of RTCN models for dimensionality reduction. Dropout probabilities are set to 0.5. For other competitor methods, we utilize the officially released implementations unless otherwise specified.

We thoroughly evaluate applicable models~\cite{pirsiavash2014assessing,parmar2017learning,xiang2018s3d,bielski2018pay,tang2020uncertainty} in video virality prediction and human-centric performance assessment domains on VDV benchmark.
For the C3D+SVR~\cite{parmar2017learning} model, we utilize a C3D network pre-trained on Kinetics dataset~\cite{carreira2017quo} and perform SVR with sklearn. 
We further evaluate more recent 3D-CNN models including I3D~\cite{carreira2017quo} and SlowFast~\cite{feichtenhofer2019slowfast} under this classic paradigm.
Moreover, state-of-the-art graph convolution models~\cite{yan2018spatial,li2020dynamic} for human action understanding are evaluated in comparison with PSGCN. DMGNN~\cite{li2020dynamic} utilizes multi-scale skeleton graphs in the encoding stage of human motion prediction, we take the encoder part of it to regress virality scores.
Lastly, sequential models of LSTM~\cite{hochreiter1997long} and TCN~\cite{bai2018empirical} are taken as baselines in comparison with RTCN. 
The LSTM model has two recurrent layers with the hidden size of 512.
The layer number and dimensionalities of TCN model consist with our RTCN.
In terms of model training, we use ADAM~\cite{kingma2014adam} optimizer with a batch size of 8. The learning rate is set to 0.0001 initially and decays every 10 epochs by the rate of 0.5.

\subsection{Virality Prediction Results}
\label{subsec:virality_prediction_results}

\begin{figure*}
\begin{center}
   \includegraphics[width=0.8\textwidth]{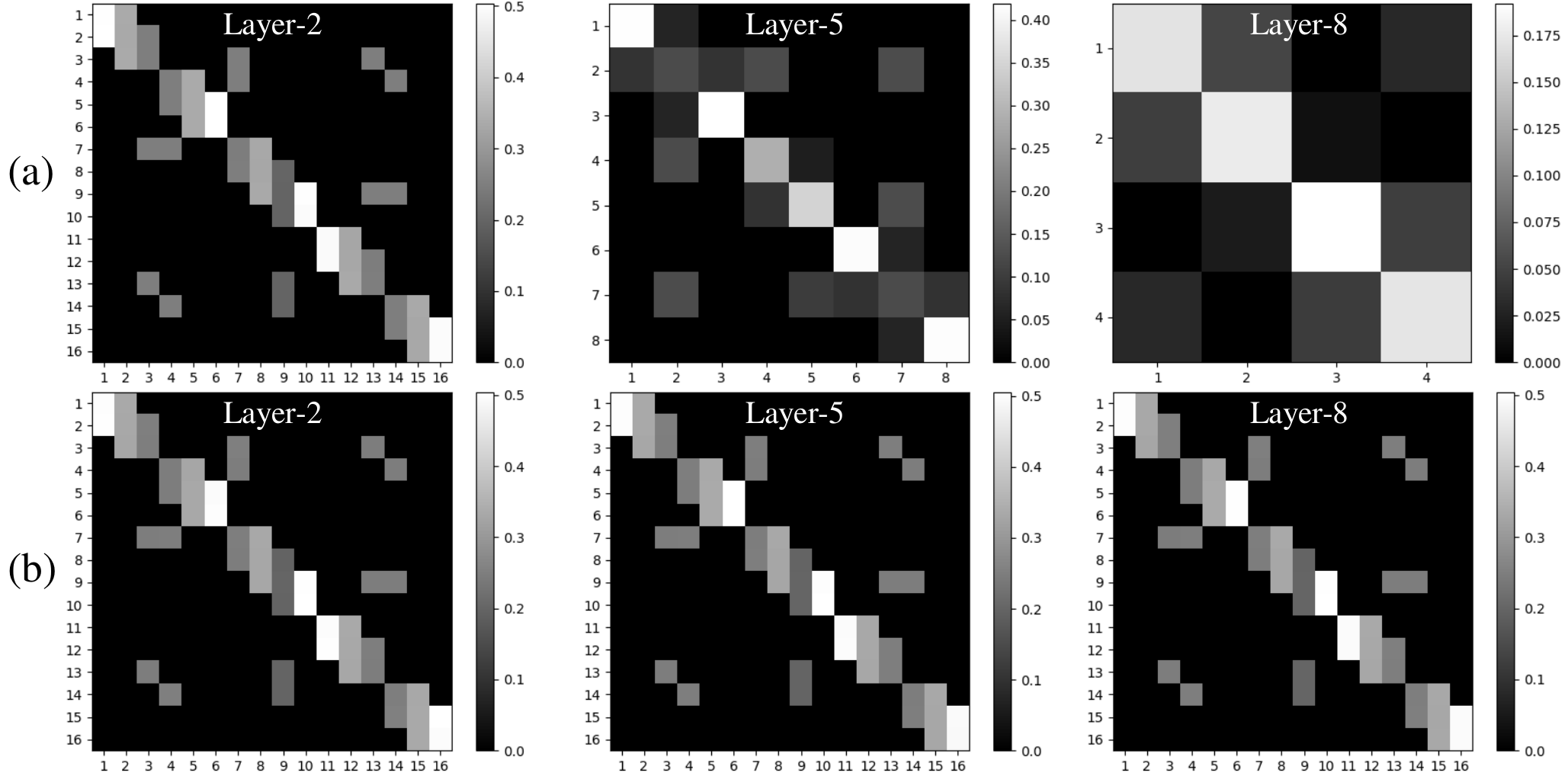}
\end{center}
    \vspace{-3mm}
   \caption{(a) Normalized adjacency matrices of skeleton graphs from the $2$-th,
   $5$-th and $8$-th graph convolution layers (from left to right) in our PSGCN. 
   Robust connections are created in upper graph layers.
   (b) Normalized adjacency matrices from the same depth in ST-GCN. 
    Graph structures remain almost the same for all layers.
}
\vspace{-6mm}
\label{fig:skeleton_graph}
\end{figure*}


\textbf{Skeleton-based prediction.} As presented in Table~\ref{tab:experiment_result}, our PSGCN shows distinct advantages over the competitors in both single- and all-challenge tracks. The Pose+DCT method has highly unstable performance across different challenges and fails in $C_1$, $C_4$, $C_6$. Compared with rigorous Olympic events, movements in viral dance are more improvisational, which makes a simple time-frequency transform like DCT inadequate to handle. In comparison with the two GCN-based models, i.e. ST-GCN and DMGNN, our method outperforms ST-GCN in all tracks and DMGNN in most tracks expect $C_1$, $C_2$, $C_3$. The advantage over ST-GCN demonstrates that the pyramidal graph architecture empowers our model with better generalization ability in virality prediction. Fig.~\ref{fig:skeleton_graph} presents examples of the skeleton graphs from the two models. It can be observed that high-level semantic connections are created in upper graph layers of PSGCN, which help to capture virality information across multiple challenges. Although DMGNN also employs multi-scale skeleton graphs, it keeps a pre-defined three-level graph pyramid in each graph convolution layer, which greatly increases the number of graph parameters. Contrastively, our PSGCN adopts a more compact architecture. We employ a single skeleton graph in each layer and gradually refine it with graph down-sampling modules. As a result, DMGNN is more prone to overfitting in tracks with limited data (e.g. $C_7$, $C_8$) compared with PSGCN.

\begin{figure*}
\begin{center}
   \includegraphics[width=0.85\textwidth]{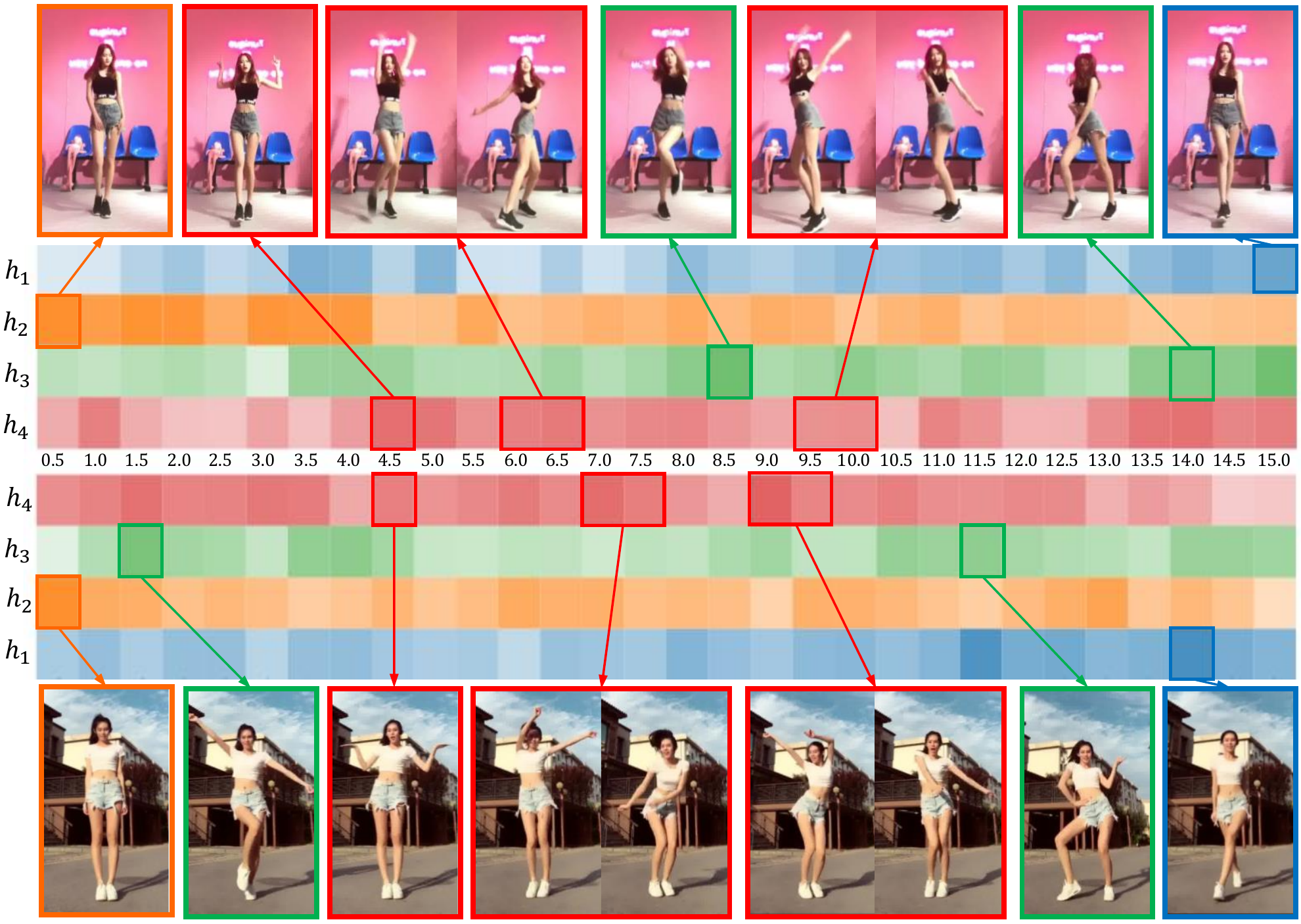}
\end{center}
    \vspace{-3mm}
   \caption{Visualization of multi-head self-attention weights in $C_7$. Attention weights from the center position of the first relational pooling block in RTCN are down-sampled into 30 bins with time-stamps given in the middle row (in seconds). Weights from the four attention heads ($h_1{\sim}h_4$) are shown in blue, orange, green and red (deeper color represents higher value), respectively. Best viewed in color.}
   \vspace{-4mm}
\label{fig:tcn_attention}
\end{figure*}

\noindent\textbf{Appearance-based prediction.} 
Since holistic features contain global appearance information of video frames, 
we utilize them to benchmark appearance-based prediction models.
As shown in Table~\ref{tab:experiment_result}, 
in single-challenge tracks, RTCN achieves competitive results in $C_3$ and $C_7$ and exceeds in other challenges. In the all-challenge track, RTCN outperforms all competitors. 
Note that although the 3D-CNN based models perform well in some challenges (e.g. S3D in $C_7$), their performance is quite unstable across all challenges. 
We also observe that more recent 3D-CNN models like I3D and SlowFast yield improvement in most cases compared with C3D.
The model in~\cite{bielski2018pay} excels in video popularity prediction tasks by utilizing spatial attention, while it does not show satisfactory results on our benchmark. This is mainly because that it averages frame-level predictions without temporal modeling, which is critical in dance virality prediction. 
Although USDL achieves better performance compared with other competitors, it is outperformed by RTCN in most tracks, suggesting that uncertainty-aware score modeling in dance virality prediction does not work as well as in sports rating.
To fairly compare the performance of RTCN and baseline sequential models, LSTM and TCN are trained with the same holistic features as RTCN. As analyzed in~\cite{bai2018empirical}, TCN has more stable gradients and better performance compared with LSTM, which is also observed on VDV benchmark. Beyond that, RTCN achieves further improvement by explicitly investigating non-local relations, outperforming the two baselines in all tracks. 

We visualize two examples of multi-head self-attention weights from $C_7$ in Fig.~\ref{fig:tcn_attention}. Attention weights from the center position of the first relational pooling block in RTCN are down-sampled into 30 bins for visualization. We observe that the four attention heads ($h_1{\sim}h_4$) attend to appearance information from different representation dimensions. Concretely, $h_1$ and $h_2$ are more likely to attend to the ending and beginning frames, respectively. $h_3$ focuses on leg movement like hop steps and bow steps, while $h_4$ pays more attention to arm movements like raising and swinging arms.

To validate the efficacy of multi-modal features in appearance-based prediction. We further incorporate facial and scenic features using attentive modality fusion.
As shown in Table~\ref{tab:experiment_result}, in most cases, adding new modalities leads to better performance, while in some cases (e.g. adding scenic features for $C_3$ and $C_5$), it degrades the performance due to the incompatibility between modalities. 
It is observed that facial features yield a more significant improvement than scenic features in the all-challenge track and most single-challenge tracks, 
indicating that facial appearance is more influential on video virality than background features on the whole. Finally, aggregating all three appearance features achieves the best all-challenge performance in appearance-based prediction, which demonstrates the efficacy of utilizing all three appearance feature streams.

\begin{figure*}[t]
\begin{center}
   \includegraphics[width=0.85\textwidth]{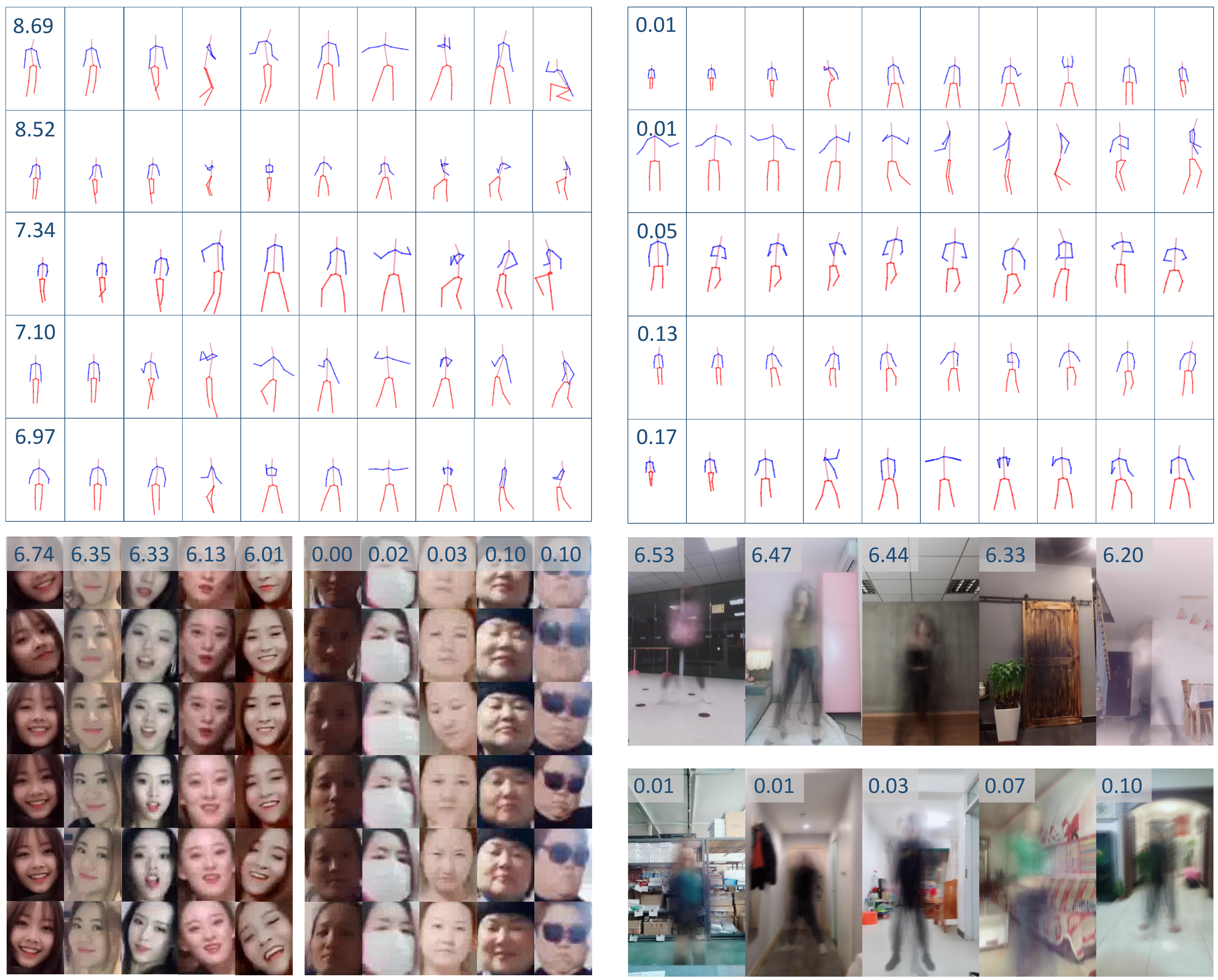}
\end{center}
    \vspace{-4mm}
   \caption{Top 5 and last 5 predicted videos from $C_1$ ranked by $s_p$, $s_f$ and $s_b$, respectively. Top: skeleton sequences ranked by $s_p$. Bottom left: facial images ranked by $s_f$. Bottom right: background images ranked by $s_b$. Multi-dimensional video retrieval and recommendation can be facilitated by this result.}
   \vspace{-6mm}
\label{fig:qualitative}
\end{figure*}

\noindent\textbf{Integrated model.}
We finally aggregate skeletal and all appearance features with attentive modality fusion in the integrated model, which is referred as \emph{Integrated} in Table~\ref{tab:experiment_result}. 
For single-challenge tracks, the integrated model outperforms other models in most challenges ($C_1$, $C_3$, $C_4$, $C_6$, $C_7$, $C_8$).
For the all-challenge track, it further improves the best SRC performance to 0.34, resulting in $30\%$ and $9\%$ improvements of the best skeleton- and appearance-based models.
This result validates the effectiveness of incorporating both body movements and appearance features in dance virality prediction, as the two modalities provide complementary information in most cases.

\noindent\textbf{Qualitative analysis.}
The proposed attentive modality fusion allows modality streams to make predictions independently. Therefore, the single modality scores produced by the integrated model, i.e. $s_p$, $s_h$, $s_f$, $s_b$, can give an qualitative explanation of what kind of videos tend to be more popular under each modality. Fig.~\ref{fig:qualitative} presents the top 5 and last 5 predicted results from $C_1$ ranked by $s_p$, $s_f$ and $s_b$, respectively. In terms of body movements, the top ranked ones have smooth motions which catch the beats, while the last ones have clumsy or even wrong movements. For example, in the 7-th frame, the top performers uniformly stretch out their limbs while the last ones show various random postures. 
For facial features, dancers with high $s_f$ scores usually have more attractive facial appearance and expressions, while the ones with low scores are less attractive and often have occlusions like masks or glasses. 
For scenic features, the background that best matches the dance style tends to have the highest score. 
For challenges with intense street dance styles like $C_1$,
dance studios (e.g. the first and third ranked backgrounds) and decorated scenes (e.g. the fourth ranked background) are preferred. This is probably because that a venue like that makes dancers look stylish and professional. 
Besides that, challenges with more gentle styles like Chinese dance also have unique background preferences.
For instance, green plants like trees and flowers are preferred in the background of $C_5$, since this dance is called ``love of fallen flowers''\footnote[1]{More details are presented in our supplementary material.}.
With only virality score supervision, our model is able to rank dance clips according to different visual aspects. 
This feature can facilitate real-world short video applications like multi-dimensional video retrieval and recommendation. Also, by averaging the attention weights of different modalities, we can provide users with suggestions about which aspect of the video matters the most for each challenge\footnotemark[1]. 

\begin{figure*}
\begin{center}
   \includegraphics[width=0.8\textwidth]{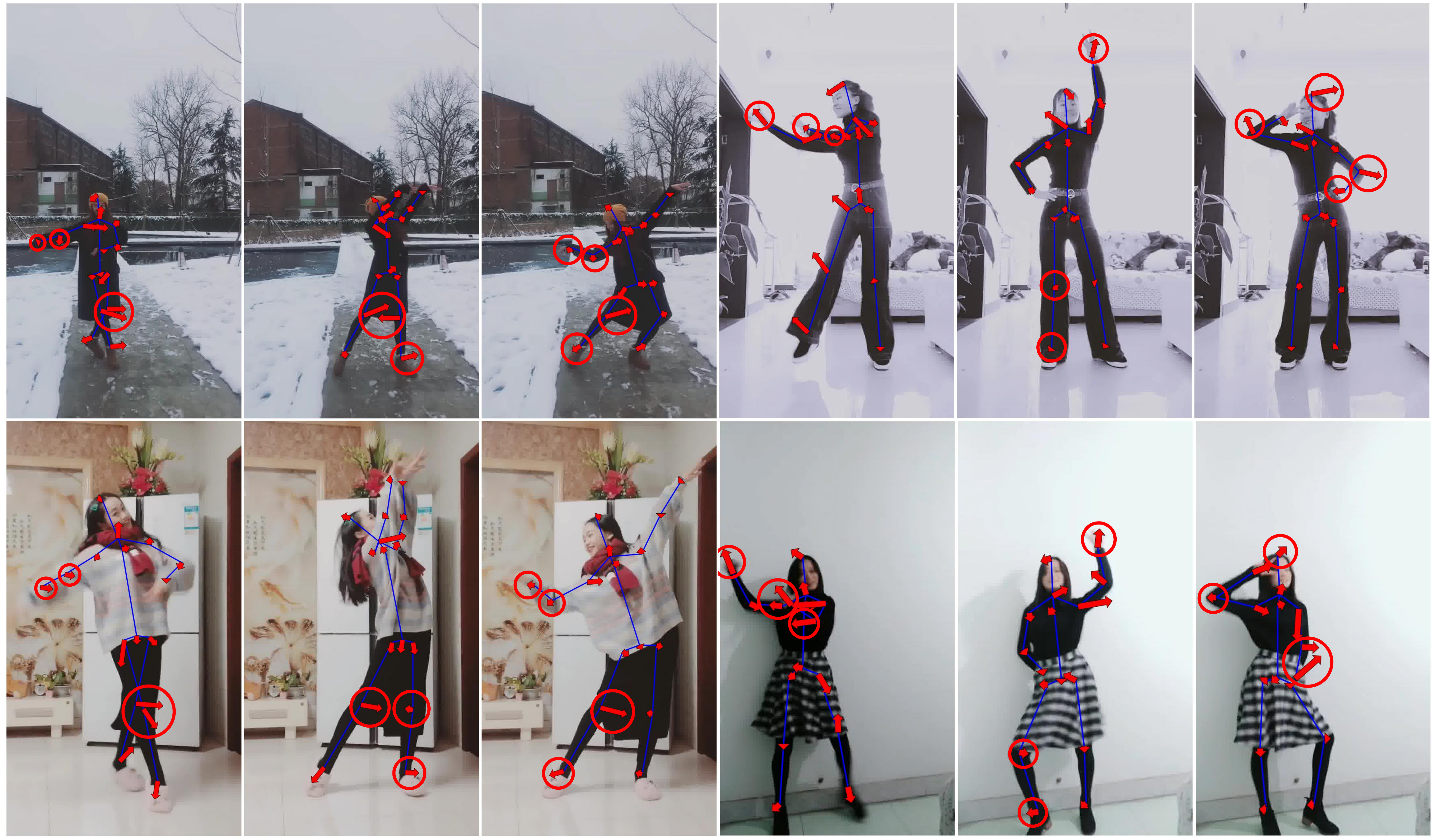}
\end{center}
    \vspace{-4mm}
   \caption{Examples of action guidance from $C_2$ (left) and $C_3$ (right). We uniformly select three critical frames in each challenge. Some important action factors are circled in red.}
   \vspace{-5mm}
\label{fig:feedback}
\end{figure*}

\begin{figure*}
\begin{center}
   \includegraphics[width=0.8\textwidth]{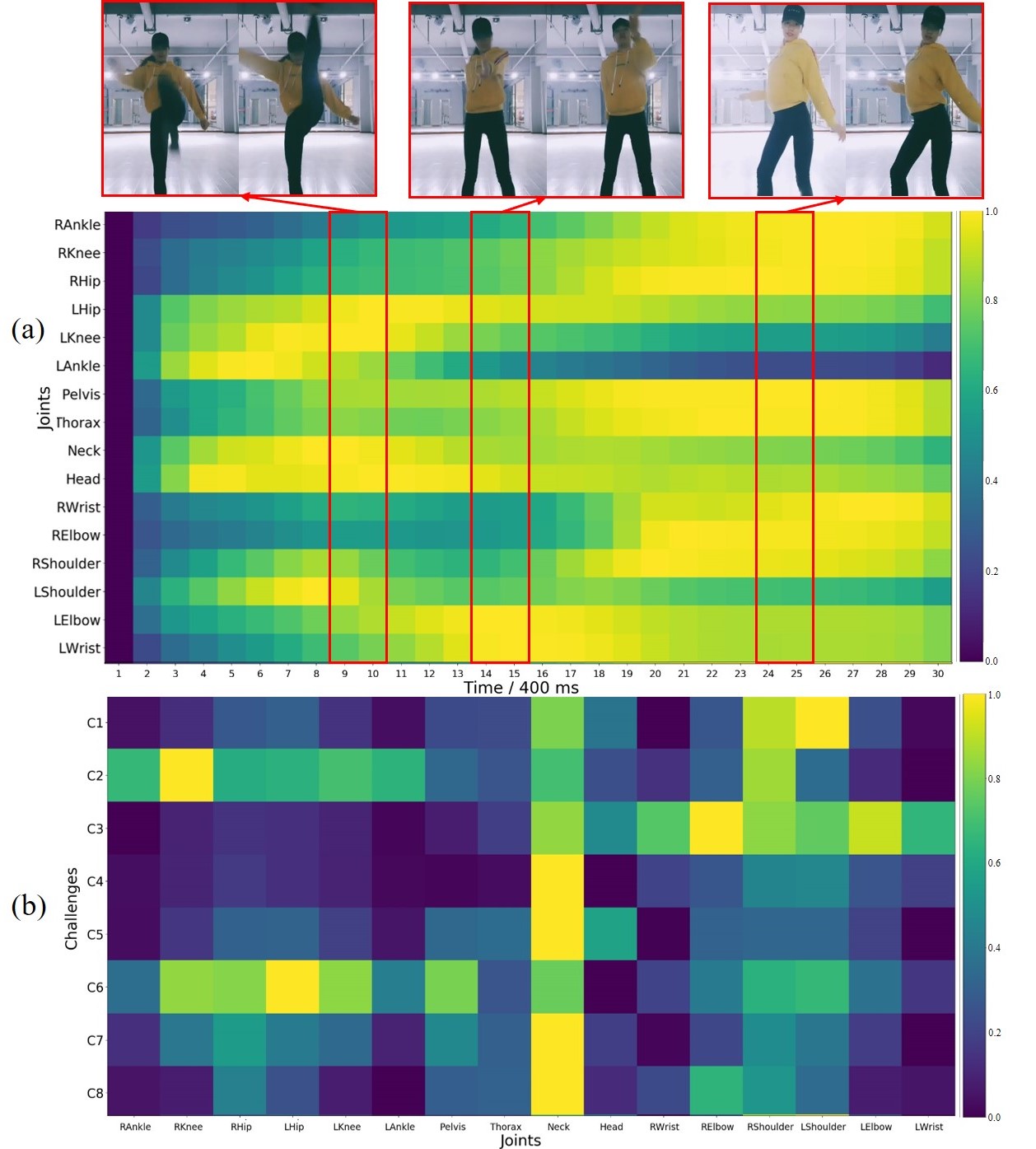}
\end{center}
    \vspace{-5mm}
   \caption{(a) Heatmap of temporal attention for challenge $C_1$. Three fragments from an example clip is shown at the top. R and L stand for right and left, respectively. (b) Heatmap of overall attention. Rows for challenges and columns for body joints. All values are normalized to $0{\sim}1$ for visualization. Best viewed in color.}
    \vspace{-6mm}
\label{fig:tvj_c1}
\end{figure*}

\subsection{Action Feedback}
\label{subsec:action_feedback}

\noindent\textbf{Action guidance.}
As introduced in Section~\ref{subsec:action_feedback_intro}, our framework generates instructive action guidance by calculating joint coordinate gradients. Fig.~\ref{fig:feedback} presents examples of generated action guidance from $C_2$ (left) and $C_3$ (right), in each of which three critical frames are selected to be visualized. We illustrate feedback vectors with red arrows, whose length shows the feedback magnitude. 
By learning through massive dance data, our PSGCN perceives important factors of dance virality, thus deriving personalized action guidance. 
For example, in the first frame of $C_2$, dancers should move right arms close to their bodies and adjust knees leftward. 
In the second frame, dancers should move knees inward with left feet spread out. 
In the third frame, it is suggested to extend right arms down while bending right legs inward. 
For challenge $C_3$, in the first frame, dancers should swing both arms upper right. 
In the second frame, it is suggested to raise left wrists high while keeping right legs straight. 
In the third frame, dancers should put left hands on their hips and move both elbows outward, while nodding heads leftward. 
As the gradient calculation can be inferred on-the-fly, dancers can obtain real-time action guidance in this way, which greatly facilitates dance learning.

\noindent\textbf{Dance tips.}
Fig.~\ref{fig:tvj_c1} (a) presents the temporal attention (defined in Eqn.~\eqref{eqn_feedback_2}) for challenge $C_1$.
Three fragments from an example clip is shown at the top. R and L stand for right and left, respectively. As the dancer performs different actions, the joint feedback magnitude varies as well. In the first fragment, where the dancer performs a high left leg kicking, the feedback of left hip, left knee and left ankle is significantly higher. In the second fragment, the active feedback of left elbow and left wrist corresponds to a shooting action performed by the left arm. In the third fragment, the dancer bends the right leg, lifts up hips and puffs out the chest, which explains the strong feedback of right ankle, right knee, hips, pelvis and thorax. By referring to temporal attention like this, dancers can learn when and which joint should be noticed through a viral dance.

Fig.~\ref{fig:tvj_c1} (b) shows the overall attention for every challenge as defined in Eqn.~\eqref{eqn_feedback_3}.
It can be observed that for challenge $C_2$ and $C_6$, the feedback for lower body joints, i.e. hips, knees and ankles, is significantly stronger. These challenges have rhythmic lower body movements like stepping forth and swaying hips. For challenge $C_1$, $C_3$ and $C_4$, feedback of upper body joints, i.e. shoulders, elbows and wrists, is more significant. This is because that upper body motions are more intense in these challenges, like shrugging shoulders and waving arms. We also find that the neck position is influential in all challenges, especially for those with active actions of both upper and lower body, e.g. $C_5$, $C_7$ and $C_8$. This is probably because that neck is the central joint which connects limbs and head. With these valuable insights, dancers can notice which body part should get more attention when participating in different challenges.


\begin{figure*}
\begin{center}
   \includegraphics[width=0.8\textwidth]{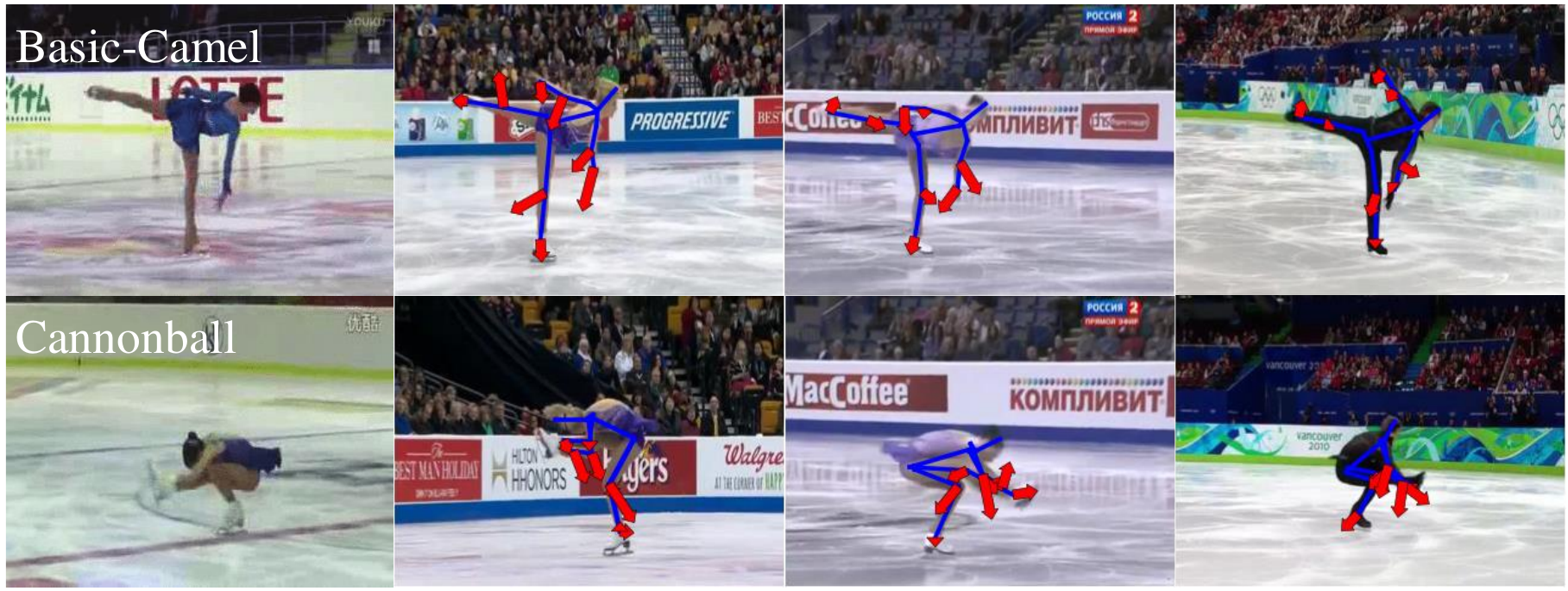}
\end{center}
    \vspace{-5mm}
   \caption{Action guidance generated for figure skating videos. Technical actions of basic-camel and cannonball from three athletes are presented with a demonstration image on the left.}
   \vspace{-6mm}
\label{fig:aqa_feedback}
\end{figure*}

\subsection{Extension to Sports Videos}
\label{subsec:results_on_AQA_tasks}
To validate the applicability of the proposed multi-modal framework in other human-centric performance assessment tasks, we extend it to sports videos and evaluate on UNLV-Sports~\cite{parmar2017learning} dataset. UNLV-Sports consists of 717 videos from three Olympic events, i.e. figure skating, gymnastic vault and platform diving. 
The performance scores in UNLV-Sports are given by expert judges according to the scoring criteria of corresponding Olympic events~\cite{parmar2017learning}. For example, a dive score is determined by the product of \emph{execution} (quality score given by judges) and \emph{difficulty} (fixed value based on dive type).
Since only skeletal and holistic features are of major concern in sports rating, we adopt these two modality streams in our multi-modal framework. Features are extracted in the same way as in VDV dataset. 

Table~\ref{tab:aqa_result} presents the comparison results  on UNLV-Sports. For skeleton-based prediction, due to the strenuous movements in Olympic events, there are many atypical body postures in this dataset, which increases the error in estimated skeletal features. Owing to the graph down-sampling modules, PSGCN is more robust to noisy inputs, thus outperforming Pose+DCT and ST-GCN by a large margin. Also, in comparison with DMGNN, the compact architecture of PSGCN shows better scalability and excels in all three events. 
For appearance-based prediction, RTCN performs the best in figure skating.
In vault and diving events, RTCN achieves competitive results compared with recent state-of-the-arts like S3D and USDL.
Finally, by integrating skeletal and holistic streams with attentive modality fusion, we observe distinct improvements in all three events. This demonstrates that our multi-modal approach is 
also beneficial to other human-centric performance assessment tasks like sports rating.

Furthermore, applications like action feedback can also be implemented in sports videos. Fig.~\ref{fig:aqa_feedback} presents examples of action guidance generated for figure skating videos. Two technical actions (i.e. basic-camel, cannonball\footnote{\url{https://en.wikipedia.org/wiki/Figure_skating_spins}}) from three athletes are shown in this figure with a demonstration image on the left. In the basic-camel posture, athletes should stretch left leg upward and move left arm close to the body. Their right arm should be dropped down straightly. In the cannonball posture, athletes should contract the skating leg downward and extend the free leg outward. The chest should be close to thighs with hands holding the ankle of the free leg. Our PSGCN learns essentials of these two actions and generates instructive action guidance.

\renewcommand\arraystretch{0.9}
\begin{table*}
\small
\caption{Results on UNLV-Sports dataset. Skeleton-based, appearance-based and multi-modal prediction results are listed from top to bottom. Best results in each experiment are shown in bold.}
\vspace{-3mm}
\begin{center}
\setlength{\tabcolsep}{1.8mm}{
\begin{tabu}{ l | c c | c  c  c}
\tabucline[1.25pt]{-}
\multirow{2}{*}{Method} & \multicolumn{2}{c|}{Feature Modalities} & \multicolumn{3}{c}{Events}\\
 & Skeletal & Holistic & Figure Skating & Gymnastic Vault & Diving \\
\hline
Pose+DCT~\cite{pirsiavash2014assessing} & \checkmark &   & 0.41 & 0.24 & 0.57 \\
ST-GCN~\cite{yan2018spatial} & \checkmark &   & 0.49 & 0.36 & 0.65 \\ 
DMGNN~\cite{li2020dynamic} & \checkmark &  & 0.51 & 0.35 & 0.67\\ 
PSGCN (ours) & \checkmark &  & \textbf{0.53} & \textbf{0.39} & \textbf{0.68} \\ 
\hline
ConvISA~\cite{le2011learning} & & \checkmark & 0.45          & -             & -             \\
C3D+LSTM~\cite{parmar2017learning} & & \checkmark & -             & 0.05          & 0.27          \\
C3D+SVR~\cite{parmar2017learning} & & \checkmark& 0.53          & 0.66          & 0.78          \\
End-to-End~\cite{li2018end}& & \checkmark & 0.58          & 0.70          & 0.80          \\
ScoringNet~\cite{li2018scoringnet} & & \checkmark & - & 0.70 & 0.84  \\
S3D~\cite{xiang2018s3d} & & \checkmark & -             & -             & \textbf{0.86} \\ 
USDL~\cite{tang2020uncertainty} & & \checkmark & 0.65 & \textbf{0.76} & 0.85 \\ 
RTCN (ours) &  & \checkmark & \textbf{0.68} & 0.74 & 0.84\\
\hline
Integrated (ours) & \checkmark & \checkmark & \textbf{0.72} & \textbf{0.76} & \textbf{0.88}\\
\tabucline[1.25pt]{-}
\end{tabu}}
\vspace{-5mm}
\end{center}
\label{tab:aqa_result}
\end{table*}

\subsection{Ablative Experiments}

\label{subsec:ablation_study}
We conduct ablative experiments to validate the effectiveness of critical components in our approach. For skeleton-based prediction, we ablate graph down-sampling modules in PSGCN, resulting in a network without pyramidal graph structures. For appearance-based prediction, relational pooling blocks are removed from the RTCN trained with holistic features. For multi-modal prediction, we replace the attentive modality fusion with simple average weighting. 
Each of the appearance modality streams in integrated model is ablated to validate their efficacy.
The effectiveness of MSE and ranking loss terms are also studied. 
Results are shown in Table~\ref{tab:ablation_study}.

Noticeable performance degradation is observed in both skeleton-based and appearance-based models, demonstrating the effectiveness of PSGCN and RTCN architectures. 
For the integrated model,
the removal of modality attention module causes performance drops in most single-challenge tracks. The all-challenge result has a even more significant drop of $15\%$ (0.34 vs. 0.29), indicating that learning inter-class modality influence is important in class-agnostic scenarios.
In terms of different modality streams, the holistic one yields the most improvement, followed by the facial and scenic streams. Although there are some exceptions in single-challenge tracks, facial and scenic features contribute in most cases. In the all-challenge track, the three modalities (i.e. holistic, facial and scenic) improve performance by $21\%$, $12\%$, $9\%$, demonstrating the effectiveness of each component in our integrated model.
The ablation of MSE and ranking loss terms also results in apparent performance decline. Removing the ranking loss term causes more degradation on the whole, which shows the importance of imposing ranking order constraints. 

\vspace{-3mm}
\renewcommand\arraystretch{0.9}
\begin{table*}[h]
\small
\caption{Results of ablative experiments. Results from skeleton-based, appearance-based and multi-modal models are listed from top to bottom. Black, red and green numbers indicate the performance is \emph{retained}, \emph{\textcolor{red}{increased}} and \emph{\textcolor{green}{decreased}}, respectively.}
\vspace{-3mm}
\begin{center}
\begin{tabu}{ l | c  c  c  c  c  c  c  c | c}
\tabucline[1.25pt]{-}
\multirow{2}{*}{Method} & \multicolumn{8}{c|}{Single-Challenge} & 
\multirow{2}{*}{All-Challenge}\\
 & $C_1$ & $C_2$  & $C_3$  & $C_4$  & $C_5$  & $C_6$  & $C_7$ & $C_8$ & \\
\hline
PSGCN & 0.25 & 0.15 & 0.22 & 0.42 & 0.33 & 0.30 & 0.51 & 0.41 & 0.26\\
w/o pyramid & \textcolor{green}{0.22} & \textcolor{green}{0.13} & \textcolor{green}{0.21} & \textcolor{green}{0.40} & \textcolor{green}{0.31} & \textcolor{green}{0.26} & \textcolor{green}{0.48} & \textcolor{green}{0.39} & \textcolor{green}{0.23}\\
\hline
RTCN & 0.29 & 0.17 & 0.24 & 0.39 & 0.20 & 0.33 & 0.42 & 0.44 & 0.27\\
w/o self-attention & \textcolor{green}{0.24} & \textcolor{green}{0.15} & \textcolor{green}{0.20} & \textcolor{green}{0.36} & \textcolor{green}{0.17} & \textcolor{green}{0.30} & \textcolor{green}{0.38} & \textcolor{green}{0.41} & \textcolor{green}{0.25} \\
\hline
Integrated & 0.37 & 0.24 & 0.30 & 0.47 & 0.33 & 0.48 & 0.54 & 0.53 & 0.34\\ 
w/o attentive fusion & \textcolor{green}{0.34} & \textcolor{red}{0.25} & \textcolor{green}{0.28} & \textcolor{green}{0.43} & \textcolor{green}{0.30} & \textcolor{green}{0.44} & \textcolor{green}{0.51} & \textcolor{green}{0.49} & \textcolor{green}{0.29}\\ 
w/o holistic stream & \textcolor{green}{0.26} & \textcolor{green}{0.20} & \textcolor{green}{0.21} & \textcolor{green}{0.44} & \textcolor{green}{0.32} & \textcolor{green}{0.37} & \textcolor{green}{0.51} & \textcolor{green}{0.46} & \textcolor{green}{0.27}\\
w/o facial stream & \textcolor{red}{0.38} & \textcolor{green}{0.19} & 0.30 & \textcolor{green}{0.45} & \textcolor{green}{0.29} & \textcolor{green}{0.45} & \textcolor{green}{0.46} & \textcolor{green}{0.48} & \textcolor{green}{0.30}\\ 
w/o scenic stream & \textcolor{green}{0.35} & \textcolor{green}{0.21} & \textcolor{green}{0.29} & \textcolor{green}{0.43} & 0.33 & \textcolor{red}{0.50} & \textcolor{green}{0.52} & \textcolor{green}{0.46} & \textcolor{green}{0.31}\\ 
w/o MSE loss     & \textcolor{green}{0.35} & \textcolor{green}{0.22} & \textcolor{green}{0.27} & \textcolor{green}{0.44} & \textcolor{green}{0.31} & \textcolor{green}{0.47} & \textcolor{green}{0.52} & \textcolor{green}{0.52} & \textcolor{green}{0.31}\\ 
w/o ranking loss & \textcolor{green}{0.32} & \textcolor{green}{0.21} & \textcolor{green}{0.24} & \textcolor{green}{0.43} & \textcolor{green}{0.28} & \textcolor{green}{0.43} & \textcolor{green}{0.50} & \textcolor{green}{0.51} & \textcolor{green}{0.28}\\ 
\tabucline[1.25pt]{-}
\end{tabu}
\vspace{-5mm}
\end{center}
\label{tab:ablation_study}
\end{table*}
\vspace{-1mm}

\section{Conclusion}
\label{sec:conclusion}
In this paper, we introduce the problem of virality prediction from dance challenges, which is of both academical and commercial value. To expedite related research, we release the VDV dataset, a TikTok based large-scale multi-modal virality prediction benchmark. We then propose a multi-modal prediction framework modeling human skeleton and appearance information with independent modality streams. For skeleton-based prediction, we devise PSGCN to hierarchically refine human skeleton graphs. For appearance-based prediction, RTCN is introduced to capture non-local temporal relations. Predictions from multi-modal streams are fused by the proposed attentive modality fusion method. Extensive experiments on VDV benchmark demonstrate the superiority of our approach. We demonstrate that practical applications like multi-dimensional video recommendation and action feedback can be facilitated by our model. Further experiments on UNLV-Sports dataset show that our multi-modal framework is beneficial to other human-centric performance assessment tasks like sports rating. 
As the potential of VDV dataset has not been fully explored yet, we will leave other promising directions like video comment generation and action retargeting to our future work.
We believe that our study will pave the way for the short video community and beyond.

\vspace{-2mm}
\begin{acks}
This work was supported by National Natural Science Foundation of China (Grant No. U20B2069), Foundation for Innovative Research Groups through the National Natural Science Foundation of China (Grant No. 61421003) and CCF-Tencent Rhino-Bird Research Fund.
\end{acks}

\bibliographystyle{ACM-Reference-Format}
\bibliography{sample-base}

\end{document}